%% file: main.tex
\documentclass[xcolor=tex,dvipsnames]{article} %
\usepackage{iclr2024_conference,times}

\input{math_commands.tex}

\usepackage{hyperref}
\usepackage{url}

\usepackage{graphicx}

\usepackage{wrapfig}
\usepackage{placeins}

\usepackage[utf8]{inputenc} %
\usepackage[T1]{fontenc}    %
\usepackage{hyperref}       %
\usepackage{url}            %
\usepackage{booktabs}       %
\usepackage{amsfonts}       %
\usepackage{nicefrac}       %
\usepackage{microtype}      %
\usepackage{makecell}
\usepackage{enumitem}

\usepackage[utf8]{inputenc}

\usepackage{txfonts}
\usepackage{bbm}
\usepackage{subcaption}
\usepackage{float}

\usepackage{multirow}

\usepackage{tikz}

\newcommand{\thickhline}{\noalign{\hrule height 2pt}}

\title{SynerGPT: In-Context Learning for Personalized Drug Synergy Prediction and Drug Design}

\author{
Carl Edwards \\
~Department of Computer Science\\
~University of Illinois Urbana-Champaign\\
~\texttt{cne2@illinois.edu} \\
\And
\quad Aakanksha Naik \\
\quad Allen Institute for Artificial Intelligence \\
\quad \texttt{aakankshan@allenai.org} \\
\And
~Tushar Khot \\
~Allen Institute for Artificial Intelligence \\
~\texttt{tushark@allenai.org} \\
\And
\quad\quad~~~~~~~~~~~~~Martin D. Burke \\
\quad\quad~~~~~~~~~~~~~Department of Chemistry \\
\quad\quad~~~~~~~~~~~~~University of Illinois Urbana-Champaign \\
\quad\quad~~~~~~~~~~~~~\texttt{mdburke@illinois.edu} \\
\And
~Heng Ji \\
~Department of Computer Science \\
~University of Illinois Urbana-Champaign \\
~\texttt{hengji@illinois.edu} \\
\And
\quad\quad\quad\quad~ Tom Hope \\
\quad\quad\quad\quad~ The Hebrew University of Jerusalem \\
\quad\quad\quad\quad~ Allen Institute for Artificial Intelligence \\
\quad\quad\quad\quad~ \texttt{tomh@allenai.org} \\
}

\iclrfinalcopy %
\begin{document}

\maketitle

\input{0abstract}

\input{1introduction}

\input{2problem_setting}

\input{3methodology}

\input{4results}

\input{5related}

\input{6conclusion}

\input{7limitations}

\clearpage

\bibliography{custom}
\bibliographystyle{iclr2024_conference}

\clearpage

\appendix

\FloatBarrier
\input{appendix/full_results}
\FloatBarrier

\input{appendix/optimization}
\FloatBarrier
\input{appendix/GA_change_over_time}
\FloatBarrier
\input{appendix/inverse_design}

\FloatBarrier
\input{appendix/tissue_type}
\FloatBarrier

\input{appendix/scaling}

\FloatBarrier

\input{appendix/context_optimization}
\FloatBarrier
\input{appendix/compute}

\FloatBarrier
\input{appendix/metrics}
\FloatBarrier
\input{appendix/drugcomb_experiment}

\end{document}

%% file: math_commands.tex
\usepackage{amsmath,amsfonts,bm}

\def\eqref#1{equation~\ref{#1}}

\def\1{\bm{1}}

\DeclareMathAlphabet{\mathsfit}{\encodingdefault}{\sfdefault}{m}{sl}
\SetMathAlphabet{\mathsfit}{bold}{\encodingdefault}{\sfdefault}{bx}{n}

%% file: 0abstract.tex
\begin{abstract}
Predicting synergistic drug combinations can help accelerate discovery of cancer treatments, particularly therapies personalized to a patient's specific tumor via biopsied cells. 
In this paper, we propose a novel setting and models for \emph{in-context drug synergy learning}. We are given a small ``personalized dataset'' of 10-20 drug synergy relationships in the context of specific cancer cell targets. Our goal is to predict additional drug synergy relationships in that context. Inspired by recent work that pre-trains a GPT language model (LM) to ``in-context learn'' common function classes, we devise novel pre-training schemes that enable a GPT model to in-context learn ``drug synergy functions''. Our model---which does not use any textual corpora, molecular fingerprints, protein interaction or any other domain-specific knowledge--- is able to achieve competitive results. We further integrate our in-context approach with a genetic algorithm to optimize model prompts and select synergy candidates to test after conducting a patient biopsy. Finally, we explore a novel task of inverse drug design which can potentially enable the design of drugs that synergize specifically to target a given patient's ``personalized dataset''. Our findings can potentially have an important impact on precision cancer medicine, and also raise intriguing questions on non-textual pre-training for LMs.\footnote{Code will be made available upon publication.}

\end{abstract}

%% file: 1introduction.tex
\section{Introduction}

Drug combination therapy is a standard practice for diseases including cancer \citep{mokhtari2017combination} and HIV. 
It is based on identifying multiple single agent therapies that, when used together, lead to synergistic effects. %
Predicting such combinatorial synergies is challenging, especially given the wide range of multiple different mutations as well as different genetic backgrounds typically found in different patients' cancer cells \citep{mroz2017challenges}. 
Many drug combinations can also cause increased toxicity \citep{zapata2020risk, juurlink2004rates} in a manner that may depend on specific patient backgrounds \citep{o2009cancer}, 
adding further complexity to the problem. 
To enable the safest and most effective implementation of combination therapy in cancer care, it is thus important to \emph{personalize} the prediction of drug synergies.%

Since the number of drug combinations scales exponentially, differentiating between synergistic and antagonistic pairings is very expensive to test in large quantities in laboratory conditions. Thus, considerable interest has recently grown in using machine learning for predicting synergistic and antagonistic effects between pairs of drugs in silico \citep{liu2020drugcombdb, preuer2018deepsynergy, rozemberczki2022moomin}. These approaches are typically not evaluated in the few-shot setting, where only a few training examples are given. This is particularly relevant in the personalized setting described above, and more generally for cancer tissue types for which there is limited training data for synergy learning models.
Additionally, these efforts use a variety of features to categorize the drugs, from molecular fingerprints \citep{preuer2018deepsynergy} to protein interactions \citep{yang2021graphsynergy}. Obtaining these features often requires integrating external knowledge sources (e.g., from drug databases), which often results in findings being restricted to the limited subsets of drugs for which this information is available and also requires specialized engineering in model design. Finally, it is unclear if these external sources are actually needed for current models.%

In this work, we address these limitations by exploring the ability of transformer language models (LMs) to learn drug synergy relations. We devise approaches that leverage transformers (1) \emph{without} any external knowledge required to be integrated into the model (i.e., no protein interaction networks or patient cell line features); (2) in the few-shot setting with an in-context learning approach that can generalize to novel unseen drugs and patient cell lines; and (3) for designing novel synergistic drug structures in the context of a specific patient's data. 

\paragraph{Transformer LMs are Strong Drug Synergy Learners---Even Without Textual Representations} First, we consider drug synergy prediction using transformer language models without enriching drugs/cells with information from external knowledge bases. We find these ``feature-less'' models are able to achieve results that are better or competitive in comparison to knowledge-enhanced state-of-art drug synergy models (e.g., BERT models achieve 84.1\% ROC-AUC to GraphSynergy's 83.4\%) %
Furthermore, in contrast to recent work that uses language models pre-trained on scientific corpora \citep{nadkarni2021scientific}, we discover an intriguing finding: using \emph{randomized} (i.e. uninformative) tokens instead of drug/cell names is able to rival models that use textual names of those entities. This suggests that external information coming from pre-training on scientific corpora (e.g., as in SciBERT \citep{beltagy2019scibert})  or the web (e.g., Wikipedia) has negligible impact on fine tuned models in this setting. These findings motivate us to explore the power of transformer models without external information, and to study generalization beyond memorization capacity by evaluating on novel drugs/cells that were unseen during training.  %

\paragraph{SynerGPT: A New In-Context Drug Synergy Setting \& Model}  We take inspiration from recent work \citep{garg2022can} that showed how a GPT model architecture can be trained to ``in-context learn'' function classes such as linear functions (e.g., linear regression/classification) and neural networks. We pre-train a GPT model from scratch on known drug synergies---using no textual corpora---and explore its ability to generalize in the few-shot setting to drugs and patient cell lines \emph{unseen} during training. We find that our model, dubbed \texttt{SynerGPT}, is able to achieve strong competitive results \emph{without} any external knowledge sources.
In particular, we introduce a new setting of \emph{In-Context Learning for Drug Synergy} (ICL-DS). In-Context Learning (ICL) \citep{dong2022survey} has emerged as a powerful paradigm for few-shot learning \citep{brown2020language}. In ICL, trained model parameters are never explicitly updated after pre-training, and adaptation to each task is done on the fly given contextual examples. This is particularly appealing in settings where it is prohibitively costly to perform parameter updates for each incoming new task and context (e.g., for each new patient in a hospital setting). We devise novel pre-training approaches for ICL-DS, including strategies for optimizing the language model prompt selection with a genetic algorithm. Prompts comprise specific combinations of drugs tested for synergy on specific patient cell lines; optimizing prompt selection in this setting has potential implications for the design of a standardized assay panel of drugs and cells to be tested for %
a patient's particular tumor. While specific patient data at this level is not readily available, we re-purpose existing drug combination data to lay the foundations for formalizing and studying our approaches from a machine learning perspective.

\paragraph{Designing New Molecules to be Synergistic in the Context of a Specific Patient} Finally, in our third major contribution we propose an additional new task of \emph{Inverse Synergistic Drug Structure Design} (ISDSD): using a GPT transformer model for \emph{generating} or \emph{retrieving} drug molecules that are synergistic in the context of a specific cancer patient's information (i.e., molecules that are synergistic with other drugs administered to a patient with specific cancer cells). This approach may in the future provide a new methodology for personalized drug candidate discovery.

%% file: 2problem_setting.tex
\section{Background and Problem Setting} \label{sec:problem_setting}

In the last few decades, combination therapy has emerged as an effective method to target genetically unstable diseases \citep{mokhtari2017combination}, with dramatic success in treating HIV \citep{moore1999natural} and more recently HCV\citep{liang2013current}. %
Unlike HIV and HCV which encode only 10-15 proteins \citep{frankel1998hiv, dubuisson2007hepatitis}, cancer is radically more complex. Since cancer has an unstable genome, combination therapy is often considered necessary \citep{mokhtari2017combination} and is commonly used in practice, with varying degrees of success.

Generally, drugs work by affecting cellular pathways--chain interactions of molecules which lead to changes in a cell. %
In \emph{drug synergy prediction}, our goal is to predict whether combining drugs will have positive or negative outcomes in the complex system of these interacting pathways. Generally, synergy lab experiments are conducted in cell lines, which are a population of cells from a multi-cellular organism (for example, human lung cancer cells). In this work, we also investigate \emph{inverse design} of drug molecules. Traditionally, the idea behind inverse design of molecules is to predict or retrieve a molecular structure which has some desired chemical property or protein target \citep{sanchez2018inverse}. In our work, we seek to explore inverse design at a higher level-- the ``interactome'' of drug interactions in complex cellular pathways.

\paragraph{General Problem Formulation} Given $k$ input drugs $d^1, d^2, \ldots, d^k \in \mathcal{D}$  along with a cell line $c \in \mathcal{C}$, the goal of drug synergy prediction is to predict a synergy value $y$ for the interactions between the drugs in the given cell line. In existing datasets, only the pairwise $k=2$ setting is considered. Thus, we focus our experiments on pairwise drug synergy, the most commonly researched setting, but our methods can naturally be extended to n-ary synergies. This problem can be considered as either a regression ($y \in \mathbb{R}$) or a binary classification problem (synergistic (True) or not (False); $y \in [0,1]$). Synergy data comes from a dataset of tuples $(d^1, d^2, c, y) \in \mathfrak{D}$. 

\paragraph{Few-Shot In-Context Setting} We also consider the few-shot setting in our formulation, which has applications for predicting synergies when there is scarce training data such as in tumor-specific synergy prediction, uncommon cancer tissues, or newly introduced single-agent therapies. %
In the few-shot setting, we assume there are $n$ synergy tuples available which contain an unknown entity $h$ (unknown cell line $c^h$ or unknown drug $d^h$). Define these tuples as $x_i := (d^1, d^2, c, y)_i$ for $i \in [1..n]$ where one of $d^1$, $d^2$, or $c$ is the unknown $h$. Each $x_i$ can then be used for training in addition to the existing training data. In our proposed method SynerGPT, we don't use these tuples $x_i$ in training-- rather, we use them as the prompt for in-context learning. Here, we are particularly interested in synergy prediction based on extremely small datasets (e.g. tested synergies from a patient's specific cancer cells), which makes traditional supervised approaches less effective. In section \ref{method:training_strategy}, we detail our training strategies for in-context learning with unknown $h$ from limited examples. 

\begin{figure*}[h]
\vspace{-.3cm}
\centering
\includegraphics[width=\linewidth]{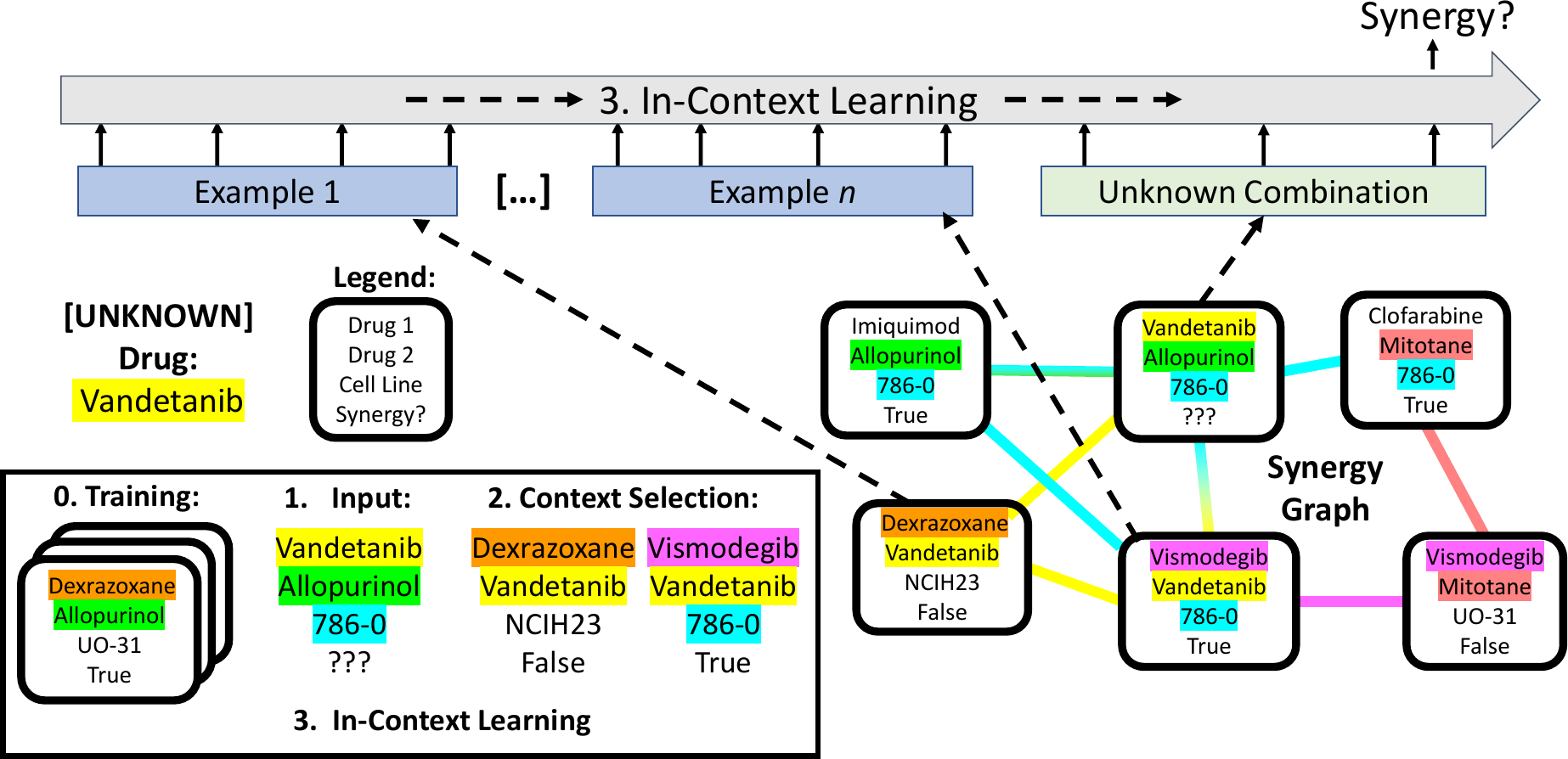}
\vspace{-.2cm}
\caption{
Example of our prompt selection strategy (steps shown in black box). After training (step 0), we are given as input (step 1) a combination between three entities: a known drug \colorbox{green}{Allopurinol}, unknown drug\colorbox{yellow}{Vandetanib}, and a known patient cell line \colorbox{SkyBlue}{786-0}. We are given a small synergy graph $\mathcal{G}$ (bottom right). Nodes represent (drug, drug, cell line) tuples with synergy labels (from previous experiments). Edges represent shared entities; edge color indicates which entities (e.g.\colorbox{pink}{red}, for sharing\colorbox{pink}{Mitotane}). Using different strategies, we adaptively select contextual examples (step 2) for in-context learning (step 3).}
\label{fig:incontext_graph}
\end{figure*}

\vspace{-.3cm}
\paragraph{Inverse Drug Design from Drug Synergy Context} %
We propose a new task where the goal is to predict the structure of a molecule given a context of drug synergy tuples (e.g., we might be given 20 synergy tuples). We train a model to predict the structure of some unknown drug $d^h$ from its synergy relations with other drugs. This has two important uses. First, this may enable scientists to predict new molecules which have desirable synergies or similar synergies to existing drugs, which is a novel way to consider drug discovery. This can potentially enable the design of drugs that synergize specifically to target a given patient's unique cancer cells. Secondly, this can support explainability of the synergy prediction model as a function of the context it is fed, by ``visualizing'' SynerGPT's understanding of the unknown drug given the context. Section \ref{sec:inverse_design} shows that we can observe the structure of the drug evolving towards the ground truth as more context examples are given. As this is a novel and difficult problem; we initially frame it as a retrieval task, effectively constraining the output space, though from an implementation perspective it is trivial to instead predict structures by using a pretrained generative model for molecules \citep{jin2020hierarchical} with no architectural differences, as both the retrieval and generation of drug structures requires generating a latent vector.

%% file: 3methodology.tex
\section{Methodology}
In this section, we will consider the four components of our paper. First, we detail how drug synergy tuples are input to encoder-only language models (§~\ref{method:encoder}). Next, we extend this idea to the few-shot ICL setting and propose training methodologies to do so (§~\ref{method:synergpt}). We then discuss optimization of the ``prompt'' used for ICL (§~\ref{method:training_strategy}). Finally, we extend our methodology to inverse drug design (§~\ref{method:inverse_design}).  

\subsection{Input for encoder-only language models}
\label{method:encoder}
Initially, we explore the efficacy of BERT-style language models \citep{devlin2019bert, beltagy2019scibert, yasunaga2022linkbert} for drug synergy prediction. We modify the task input to be in natural language using a simple formulation:\\
\centerline{[CLS] $d^1$ [SEP] $d^2$ [SEP] $c$ [SEP]} \\
where $d^1$ and $d^2$ are drug names (e.g., \emph{imatinib}, \emph{5-FU}), 
and $c$ is the name of a cell line (e.g., \emph{MCF2}, \emph{Ishikawa}). The model is then trained to predict the output value $y$ from the \textbf{[CLS]} token representation. 

We also investigate to what extent pretraining knowledge is responsible for the model's performance. To do so, we evaluate the impact on performance when the drug and cell names are replaced with `random' tokens. Given the ordered (by frequency) %
vocabulary $\mathcal{V}$ of the LM, we select the tokens $\{v_i \in \mathcal{V} \mid i \in \left[k..(k+|\mathcal{C}|+|\mathcal{D}|)\right]\}$ to represent our drug and cell lines. Note we start at a threshold $k$ to avoid the most common tokens which might have specialized representations in the language model's latent space. %
We uniquely map each cell line and drug to a token in this set, which we use as input to the BERT LM. Essentially, this experiment is used to determine whether knowledge from pretraining or the transformer architecture itself is responsible for performance on the drug synergy task.
An example input from this strategy is: {[CLS] rabbit [SEP] fish [SEP] book [SEP]}.

\subsection{SynerGPT: In-Context Learning for Few-Shot Synergy Prediction}
\label{method:synergpt}

\subsubsection{In-Context Learning for Function Classes: Background}
Recent work trained transformer models to ``in-context learn'' function classes \citep{garg2022can}. A function class is a set of functions that satisfy specific properties, such as linear functions or neural networks. In-context learning of a function class $\mathcal{F}$ is defined as being able to approximate $f(x_\textrm{query})$ for ``most'' functions $f \in \mathcal{F}$ given a new query $x_\textrm{query}$ when conditioned on a prompt sequence $(x_1, f(x_1), \ldots, x_n, f(x_n), x_\textrm{query})$. We define a prompt prefix $P^n \coloneqq (x_1, f(x_1), \ldots, x_n, f(x_n), x_{n+1})$ as the first $n$ in-context examples followed by the $n+1$th input. A model $M_\theta$ parameterized by $\theta$ is trained to minimize the loss averaged over all prefixes
\begin{equation} 
\min_\theta \mathbb{E} \left[ \frac{1}{n+1} \sum_{i=0}^n w_i \ell \left( M_\theta (P^i), f(x_{i+1})\right) \right] 
\label{training_eq}
\end{equation}
given some appropriate loss function $\ell$. Weights $w_n := 1$ unless otherwise noted. %

\subsubsection{Predicting Drug Synergy In-Context} \label{method:incontext}
For in-context prediction of drug synergy, we redefine 
$$P^n = (d_1^1, d_1^2, c_1, y_1, \ldots, d_n^1, d_n^2, c_n, y_n, d_{n+1}^1, d_{n+1}^2, c_{n+1})$$
as the prompt prefix (as discussed in Section \ref{sec:problem_setting}, we refer to this as the ``context'' or ``input context''). Here, $y$ can be considered the output of a function measuring synergy on $(d^1, d^2, c)$. %
As in \citep{garg2022can}, we consider a GPT-2 family \citep{radford2019language} decoder-only model, which we call SynerGPT. Here, the prediction of the synergy value $y_{j}$ is made using a linear transformation of the contextualized output representation of $c_{j}$ (note that this includes $d_j^1$ and $d_j^2$ due to self-attention). %
Model inputs--drugs $d$, cell lines $c$, and labels $y$--are initialized using a learnable embedding layer (i.e. no external features). 
To evaluate the model's ability to predict synergies of unknown entities, we hold out either $m$ drugs or $m$ cell lines and remove their synergy relations from the training set (see Section \ref{sec:results}). We use a subset of the held out tuples as a pool of context examples. %
We now turn to selecting the context (prompt prefix) from this pool in a manner that increases predictive performance. %

\subsubsection{How to sample the context?}
\label{method:training_strategy}

A central question about using language models without external features---including textual names---is how to teach the model to understand unknown drugs or cell lines. We propose using a masking strategy---every unknown drug $d^h$ or cell $c^h$ is represented by \textbf{[UNKNOWN]} and the model must use in-context learning to understand it based on contextually-related known drugs and cell lines. In this setting, we assume that we are given a set of synergy tuples to sample from to construct a prompt. During training, it's simply the training set. During evaluation, we consider a special held-out ``context'' set $\mathfrak{D}^c \subset \mathfrak{D}$ (thus named because we sample the context/prompt $P^n$ from this set). To sample from this context set, we propose a context-selection strategy based on constructing a graph $\mathcal{G}$ on this $\mathfrak{D}^c$. %
Specifically, we construct $\mathcal{G}$ by creating a node for every synergy tuple $x := (d^1, d^2, c, y) \in \mathfrak{D}^c$. We construct a drug edge $e^d$ between two nodes $x_1$ and $x_2$ if they share drug $d$ (i.e. $d \in x_1 \wedge d \in x_2$). Similarly, we construct a cell line %
edge $e^c$ if they share cell line $c$. See Figure \ref{fig:incontext_graph} for an example and Appendix Figure \ref{fig:incontext_archi} for more details. 
We employ the following context selection strategies %
to sample a context with $n$ examples given some node $x$ containing unknown $h$ which is either drug $d^h$ or cell $c^h$: 

\begin{enumerate}[wide, labelwidth=!,itemindent=!,labelindent=0pt]
    \item \textbf{Random}: Uniformly select $n$ context examples from $\mathfrak{D}^c$. 
    \item \textbf{Graph}: Uniformly select examples from the nodes adjacent to $x$ in $\mathcal{G}$. %
    \item \textbf{Unknown-First}: Uniformly select nodes adjacent to $x$ which share an edge of type $e^{h}$, i.e. prioritizing selection of nodes that contain the masked unknown $h$. %
\end{enumerate}
Note that these strategies are hierarchical-- \textbf{Unknown-First} falls back to \textbf{Graph} when there aren't enough examples which falls back to \textbf{Random}. Examples from \textbf{Random} are put earlier in the context than \textbf{Graph} which is again put before \textbf{Unknown-First}. In order to train the model to correctly use the \textbf{[UNKNOWN]} token, we need to artificially create unknown drugs or cells during training. Given training example $x$, we uniformly select $d^1 \in x$ or $d^2 \in x$ to be the hidden drug $d^h$. For the unknown cell line setting, $c \in x$ is always set to $c^h$ because there is just one cell line per example. We replace all occurrences of $h$ in the prompt with \textbf{[UNKNOWN]}. We note that our sampling strategy is related to retrieval augmented models \citep{mialon2023augmented}. Here, however, we note that the model is also in-context learning synergy functions for unknown drugs based on Definition 1 in \citep{garg2022can}.

\subsubsection{Optimizing the Context} \label{method:context_optim}

We further study whether the context can be optimized to best enable predictions for some unknown drug or cell line $h$ (see Figure \ref{fig:contextoptim} for an example). The purpose of these experiments is to enable the eventual development of a standardized assay for drug synergy prediction. Thus, as output, these optimization algorithms produce a set of context tuples for each $h$. To do this optimization, we assume that we have four splits of data, which are constructed as follows. Given a set of $p$ ``unknown'' drugs/cells $H$, all synergy tuples not containing any $h \in H$ are put into a training set $\mathfrak{D}^{Tr}$. The remaining tuples are randomly partitioned into three equal sized sets: a context bank $\mathfrak{D}^c$, a validation set $\mathfrak{D}^v$, and a test set $\mathfrak{D}^{Te}$. %
We first train a model on $\mathfrak{D}^{Tr}$ following the \textbf{Unknown-First} strategy (where contexts are sampled from $\mathfrak{D}^{Tr}$ itself). Following this, for each unknown entity $h_i$, we select $n$ context examples from $\mathfrak{D}^c$ which maximize the model's score on the validation set $\mathfrak{D}^v$. This is a combinatorial optimization problem which can be considered related to the best subset selection problem \citep{bertsimas2016best, miller2002subset}. We consider a genetic algorithm \citep{gad2021pygad}: %
a metaheuristic method which is useful for black box optimization of systems containing complex interacting parts \citep{mitchell2007machine}, %
which is suitable for the complex interactions between cellular pathways required for drug synergy prediction. As output, we get a set of context tuples for each $h$. Optimization algorithm details are given in Appendix \ref{appendix:optimization}. %

\subsection{In-Context Learning for Inverse Design}
\label{method:inverse_design}

To train the model to retrieve relevant drug structures in-context, we use the same architecture as for synergy prediction (§~\ref{method:incontext}), so that we can use the same data split and optimized contexts from Section \ref{method:context_optim} to understand how the model interprets them. For effective retrieval, we need a strong base molecular representation that makes it possible to effectively distinguish molecules. %
So, we choose to use MegaMolBARTv2 \citep{megamolbartv2} representations, which were trained on 1.45 billion molecular SMILES strings and thus have a relatively comprehensive (in terms of drug classes) latent space. We train a SynerGPT model from scratch %
to predict representations using a linear transformation on the output \textbf{[UNKNOWN]} representation. We use this final representation to retrieve the desired drug using cosine similarity with the MegaMolBARTv2 representations of the drugs in our synergy dataset. The training context is selected using the \textbf{Unknown-First} strategy. Finally, we train the model using a minibatch contrastive loss \citep{radford2021learning, edwards2021text2mol} between the L2-normalized ground truth representations $D^g$ (here MegaMolBartv2) and predicted representations $D^p$ (output from our model's prediction head):
\begin{equation} \label{clip_loss}
\ell(D^g,D^p) = CE(e^\tau D^g{D^p}^T, I_b) + CE(e^\tau D^p{D^g}^T, I_b)
\end{equation}
where $CE$ is categorical cross-entropy loss, $b$ is the mini-batch size, $I_b$ is the identity matrix, and $\tau$ is a learnable temperature parameter. We use this loss for $\ell$ in equation \ref{training_eq}.

%% file: 4results.tex
\section{Results} \label{sec:results}

\paragraph{BERT can do Drug Synergy?} 
\label{results:bert}
In this section, we experiment with finetuning BERT on drug synergy data where all drugs and cell lines are seen during training (data splits detailed in Appendix~\ref{appendix:graphsynergy_results}).
As discussed earlier, %
there has been recent work using external network datasets capturing interactions between drugs, proteins and cell lines \citep{yang2021graphsynergy} for synergy prediction. %

\begin{wraptable}{r}{0.7\textwidth}
\centering
\vspace{-.5cm}
\begin{subtable}{0.7\textwidth}
\resizebox{\columnwidth}{!}{
\begin{tabular}{l|cc|cc}%

 Model & \thead{KB} & \thead{Name} & ROC-AUC & PR-AUC \\
\thickhline
 DeepSynergy & $\times$ &  & 84.3 & 70.4 \\
 \hline
 MR-GNN & $\times$ &  & 77.9 & 62.6 \\
 \hline
 SSI-DDI & $\times$ &  & 63.3 & 41.4 \\
 \hline
 DeepDDS & $\times$ & & 87.2 & 77.0 \\
 \hline
 SciBERT (random) & & & 86.9 & 76.3 \\
 \hline
 BioLinkBERT (names) & & $\times$ & 86.4 & 75.9 \\

\end{tabular}
}
\end{subtable}
\vspace{-.1cm}
\caption{
Classification results for four selected ChemicalX \citep{rozemberczki2022chemicalx} baselines and BERT on DrugCombDB \citep{liu2020drugcombdb}. SciBERT and BioLinkBERT take random token and names as input, respectively. Values are average of five runs. Notably, SciBERT (random) outperforms four of the other five baselines. KB means external knowledge is used.}
\vspace{-.4cm}
\label{tab:transductive_results}
\end{wraptable} 
To evaluate the impact of these external datasets, we compare against a strong and recent model, Graphsynergy \citep{yang2021graphsynergy} that uses over a dozen different network datasets and achieves state-of-the-art on its subset of DrugCombDB.  %

We train four BERT-based \citep{devlin2019bert} language models \citep{beltagy2019scibert, yasunaga2022linkbert} and find that they outperform GraphSynergy in both name and random token settings. %
BioLinkBERT with random tokens, for example, achieves a ROC-AUC %
score of 84.1\% compared to GraphSynergy's 83.4\% (p < 0.05 using paired $t$-test). In comparison, BioLinkBERT with drug names as input %
achieves 83.6\%. %
We checked multiple BERT configurations, and details on other BERT models are shown in Appendix \ref{appendix:graphsynergy_results} Table \ref{tab:full_graphsynergy_transductive_results}. %

A natural question here is whether the model has learnt the required knowledge during pre-training. Surprisingly, replacing drug and cell names with random tokens (§~\ref{method:encoder}) resulted in no drop in performance. This suggests that the transformer architecture may be the dominant factor explaining BERT's performance on the task. However, if we use a randomly-initialized BERT model without any pre-training, we find the performance is worse (by ~3 ROC-AUC pts). %

To verify our findings, we consider the ChemicalX framework \citep{rozemberczki2022chemicalx}, which implements several baselines and provides a standardized subset of DrugCombDB \citep{liu2020drugcombdb} with drug and cell line features. This standardization allows us to compare different baseline methodologies on the same dataset.%
The ChemicalX DrugCombDB dataset has 2,956 drugs, 112 cell lines, and 191,391 synergy tuples. We compare against baselines DeepSynergy \citep{preuer2018deepsynergy}, MR-GNN \citep{xu2019mr}, SSI-DDI \citep{nyamabo2021ssi}, and DeepDDS \citep{wang2022deepdds}, which we train using default hyperparameters from the original papers for 50 epochs as in \citep{rozemberczki2022chemicalx}. 
These baselines (details in Appendix \ref{appendix:few_shot_full_results}) represent the most popular approaches to drug synergy prediction and allow us to compare against transformer architecture performance. Remarkably, SciBERT with \emph{random} tokens outperforms all baselines except DeepDDS in this setting (Table~\ref{tab:transductive_results}). We see similar results on the DrugComb dataset (this database is larger but is continuously modified by volunteers; see Appendix \ref{appendix:drugcomb}). We note that, while this performance is surprising in this domain, it follows from results from other domains. For example, language models are able to learn complex grammar and interactions just by observing how words co-occur. We conjecture this may be related to the observation that pre-training on a nonsense corpus \citep{krishna2021does} can provide good weight initializiations for downstream tasks. We further discuss related work in Section \ref{appendix:related}.

\vspace{-.2cm}
\subsection{In-Context Learning for Few-Shot Drug Synergy}
\label{sec:icl_results}

We now evaluate models on the few-shot and zero-shot setting, i.e, when a new drug or cell line is introduced with limited or no interaction data.
We use the same architecture used in \citet{garg2022can}: a GPT-2 \citep{radford2019language} model with 256-dimensional embeddings, 12 layers, 4 attention heads, and batch size of 64. We use a learning rate of 2e-5. Model weights are initialized from scratch. To enable efficient experimentation in the few-shot setting, we construct a dataset split which contains multiple unknowns (i.e. $m$ held-out drugs or cells: $H := \{h_i \mid i \in [1..m]\}$). To construct our split, we remove all synergy tuples containing $h \in H$ from the dataset $\mathfrak{D}$ so that the remaining dataset only contains tuples with known drugs/cells (this is our training set $\mathfrak{D}^{Tr}$).
Then, for each $h$, we select $n$ synergy tuples randomly to form the ``context'' bank/split $\mathfrak{D^c}$. All other ``unknown'' synergy tuples are put into $\mathfrak{D}^{Te}$. %

For comparison, we use the same baselines trained in zero-shot and few-shot settings. We also test SetFit \citep{tunstall2022efficient} (a few-shot LM approach), k-nearest neighbors, off-the-shelf pre-trained GPT-2 (using entity names as input, similar to CancerGPT \citep{li2023cancergpt}), and MAML with DeepDDS (details in Appendix \ref{appendix:few_shot_full_results}). In the few-shot setting, the context bank $\mathfrak{D}^c$ is considered part of the training set, and in the zero-shot setting it is not used. Our model, SynerGPT, however, is not trained on the context bank but uses it as context (prompt) examples for evaluation. Examples are selected using the Random, Graph, or Unknown-First strategies. We separately investigate the setting where drugs are unknown and where cell lines are unknown.

\textbf{Unknown Drugs} %
To construct the dataset split, we set $m = 50$ unknown, i.e.,``held-out'' drugs and context $n = 20$ synergy tuples. Hence, our context bank contains $50 \times 20 = 1,000$ tuples. 
Overall, we find that our SynerGPT can perform better in the few-shot setting than existing baselines on on this task, as shown in Table \ref{tab:inductive_results}. Full results are in Appendix Table \ref{tab:full_inductive_results}. %
SynerGPT is trained in the zero-shot setting, which means it can be evaluated both with context examples (few-shot) and without any examples (zero-shot).  Each strategy performs roughly the same zero-shot (although since strategies are used in training there are small differences)%
, but the performance with sampled context examples is much different. Without examples, SynerGPT performs worse than DeepDDS few-shot, but the same SynerGPT model outperforms DeepDDS when given the few-shot context. Overall, we outperform all prior models in the few-shot setting and zero-shot setting. %
 In particular, Unknown-First is able to increase performance by 3.8\% absolute ROC-AUC with context, whereas DeepDDS only increases 1.3\% from zero- to few-shot. Our approach is able to leverage the few given examples more effectively as shown by this higher increase in ROC-AUC. %
It is also notable that Unknown-First outperforms Graph since the context contains more examples with the unknown drug which the model is able to utilize to produce better predictions. %

For example, the tuple (Vismodegib, Mithramycin A, NCI-H226) with unknown Vismodegib is True. Without examples, this is predicted as 0.46. For Graph with examples, it is predicted as 0.65--closer to the ground truth. For Unknown-First, the prediction further increases to 0.79. In this example, Graph only sees 15 examples containing the unknown but Unknown-First sees a full 20. Few-shot DeepDDS predicts 0.47 for this example, which is quite similar to our method without examples.
As another example, (Chlorambucil, Cylocide, SK-OV-3) consists of two unknown drugs and has label False. Without examples, it is predicted as 0.62. Graph improves this to 0.35 and Unknown-First improves to 0.23. Interestingly, few-shot DeepDDS exhibits high uncertainty and predicts 0.50.

\begin{table}%
\centering
\begin{subtable}{.8\textwidth}
\resizebox{\columnwidth}{!}{

\begin{tabular}{ c|l|cc|cc }

 \multicolumn{2}{c}{} & \multicolumn{2}{c}{\underline{Unknown Drug}} & \multicolumn{2}{c}{\underline{Unknown Cell Line}} \\ 
Mode & \multicolumn{1}{c}{Model} & \multicolumn{1}{c}{ROC-AUC} & PR-AUC & ROC-AUC & \multicolumn{1}{c}{PR-AUC} \\
\thickhline
 \multirow{6}{*}{Zero-Shot} & DeepSynergy & 67.5 & 47.7 & 78.6 & 63.6 \\ 
 & DeepDDS & 72.1 & 53.2 & 74.5 & 59.8 \\
 & SciBERT (random) & 67.7 & 47.4 & 79.1 & 64.4 \\
 & MAML-DeepDDS & 68.76 & 50.05 & 71.6 & 54.6 \\
 & kNN-Features & 65.4 & 45.9 & 82.0 & 70.3 \\
  & SynerGPT* (ours) & \textbf{74.0} & \textbf{57.3} & \textbf{83.5} & \textbf{72.1} \\
 \thickhline
 \multirow{8}{*}{Few-Shot} & DeepSynergy & 71.6 & 53.9 & 82.0 & 68.7 \\
 & DeepDDS & 75.5 & 57.4 & 74.2 & 60.4 \\
 & SciBERT (random) & 73.8 & 56.9 & 80.5 & 66.4 \\
 & MAML-DeepDDS & 68.79 & 50.00 & 71.4 & 54.6 \\
 & kNN-Features & 66.9 & 47.7 & 82.1 & 70.5 \\
 & SetFit-S2 & 58.8 & 39.4 & 63.3 & 44.6 \\
 & GPT-2 & 74.2 & 56.8 & 80.3 & 66.6 \\
  & SynerGPT* (ours) & \textbf{77.7} & \textbf{61.5} & \textbf{83.8} & \textbf{72.8}  \\

\end{tabular}
}
\end{subtable}
\caption{Few-shot and zero-shot results on ChemicalX DrugCombDB with 50 unknown drugs / 20 unknown cell lines. Our in-context methods perform better than baselines trained in the few-shot setting. Results are averaged over 5 runs. Zero-shot SynerGPT is evaluated without context. BERT models use random tokens. The difference between SynerGPT with and without context has p < 0.05 for both unknown drugs and cell lines based on a paired \textit{t}-test. %
Similarly, both are statistically significant from the best baseline. *For simplicity, we report the best selection strategy (Unknown-First for unknown drug and Interpolate for unknown cell line). Full results are in Appendix \ref{appendix:few_shot_full_results}.} 

\label{tab:inductive_results}
\end{table} 
\raggedbottom

\textbf{Unknown Cell Lines} Since there are only 112 cell lines, we set $m = 20$ as unknown and use $n=10$ context examples. Interestingly, we find that models perform worse with context examples. We believe this is caused by the relatively small number of patient cell lines in the data vs. 2,956 drugs, making it harder to learn higher-level types of drug-cell line interaction. In other words, we are trying to learn a complex function class (drug synergy in an unknown cell line) without a significant number of example functions $f\in \mathcal{F}$. To alleviate this issue, we use 6 layers, batch size of 128, and only 30 epochs. Nonetheless, the issue still exists--performance decreases for baselines DeepDDS and MR-GNN and our strategies Unknown-First and Graph. %
We experiment with interpolating between training initially with Random to Unknown-First at the end (see Appendix \ref{appendix:interpolate}), which helps in the unknown cell line case. We believe this creates an exploration-exploitation effect. %

\subsection{Context Optimization}
\label{sec:context_optim}

\begin{wraptable}[9]{r}{0.5\textwidth}
\centering
\vspace{-.5cm}
\begin{subtable}{0.5\textwidth}
\resizebox{\columnwidth}{!}{
\begin{tabular}{c|c|c|c|c }

 \multicolumn{1}{c}{} & \multicolumn{2}{c}{\underline{Unknown Drug}} & \multicolumn{2}{c}{\underline{Unknown Cell Line}} \\ 
 \multicolumn{1}{c}{Strategy} & \multicolumn{1}{c}{ROC-AUC} & PR-AUC & ROC-AUC & \multicolumn{1}{c}{PR-AUC} \\
\thickhline
 Mean UF & 79.2 & 63.8 & 85.2 & 74.9 \\
 \hline
 Best UF & 80.8 & 66.4 & 85.6 & 75.7 \\
 \hline
 GA & \textbf{81.5} & \textbf{66.9} & \textbf{86.1} & \textbf{76.5} \\

\end{tabular}}
\end{subtable}
\caption{Test-set context optimization results ($p < 0.0001$).
Model parameters are fixed, only context is changed. UF indicates Unknown-First strategy.}
\label{tab:optimization}
\end{wraptable}

As we have shown in the previous section that the context selection strategy is very important for SynerGPT performance, the natural next question is to what extent the context can affect model performance. To test this, we conduct a different split. Like before, we select 50 unknown drugs and 20 cell lines; with their respective tuples, we create three uniform splits: context, validation, and test. We train a SynerGPT Unknown-First model using hyperparameters as in our above experiments. 

In context optimization, our goal is to select examples from the context and train splits which maximize some metric on the validation split. For our experiments, we maximize ROC-AUC for our trained model using the validation set. Overall, we consider two strategies: Unknown-First, and a genetic algorithm (GA). %
For the genetic algorithm, we use the implementation and hyperparameters from PyGAD \citep{gad2021pygad} with a population of 8 for 50 epochs. Here, we consider each example in the context split to be a potential gene. For comparison, we also select the context at random according to the Unknown-First strategy. To ensure comparability, we evaluate Unknown-First the same number of times as the genetic algorithm and select the best context. %
Our results (Table \ref{tab:optimization}) show that the genetic algorithm optimizes the context from a starting average AUC of 79.2\% up to 81.5\% for unknown drugs and from 85.2\% to 86.1\% for unknown cells. Appendix \ref{appendix:GA_change} visualizes this and shows error bars. %
We further analyze the results by different tissue types (Appendix \ref{appendix:tissue_type}). For example, we find that for unknown drugs, synergy prediction in ovarian cancer is effective, but for both unknown drugs and cell lines predictive performance on bone cell lines is low. %

\subsection{Inverse drug design from  synergy examples for discovery and explainability}
\label{sec:inverse_design}
\begin{figure}[!t]
\centering
\includegraphics[width=\linewidth]{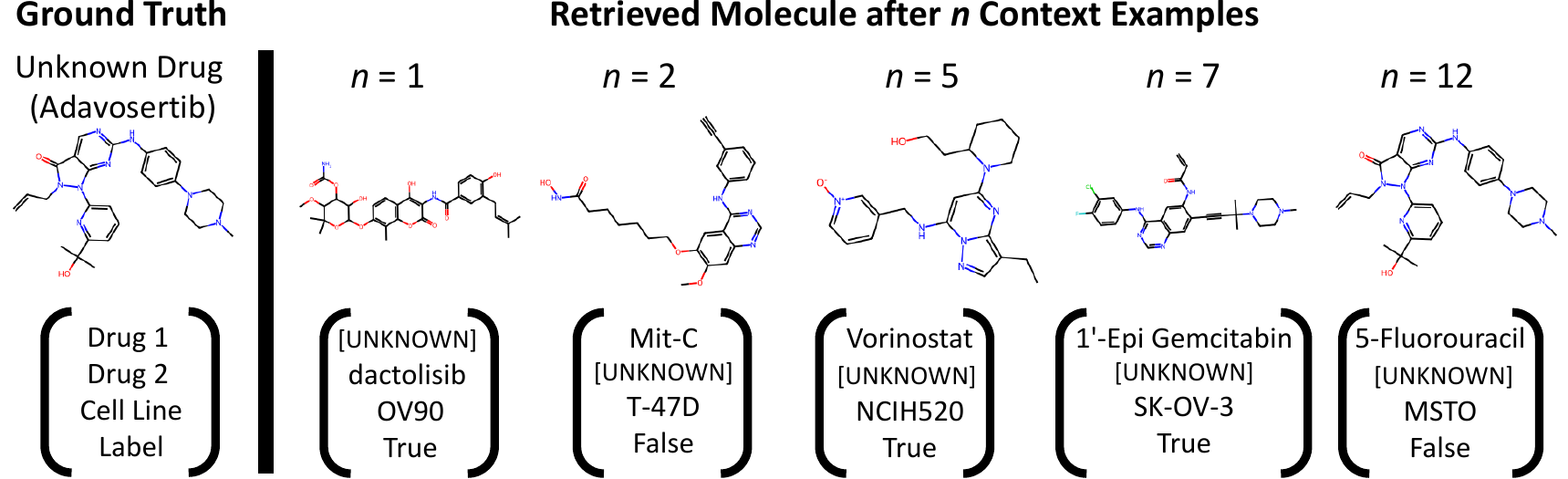}
\caption{This figure shows the model's understanding of an unknown drug (Adavosertib) by retrieving candidates from the pool of held-out drugs. As the model sees more context synergy tuples $n$ (shown at the bottom; selected by the GA), it retrieves structures closer to the target molecule until finding the ground truth after 12 context examples. Repeated structures are skipped for brevity.}
\label{fig:unseen_only_example}
\end{figure}

Explainability is one of the most challenging problems in deep learning. With transformer language models, the contrast between remarkable performance gain and lack of explainability becomes even more striking. Here we propose a novel drug design task to better understand the model's ``thought process.'' As shown in Figure \ref{fig:unseen_only_example}, essentially, we look at SynerGPT's prediction as it gains more information via synergy tuples. While this is a useful step, we do recognize that retrieval doesn't fully address explainability and hope to inspire further work. We refer to Limitations (§~\ref{limitations}) for more discussion. 

In this novel task, we evaluate SynerGPT's ability to retrieve the structure of an unknown drug. We use the same splits as before but replace the classification head with a vector output trained using the loss in Equation \ref{clip_loss}. Using the same splits allows us to visualize the optimized context from the genetic algorithm.
Experimentally, we achieve the best performance with the weight value from equation \ref{training_eq} set to $w_i := i/k$. Two examples of the model retrieving drugs which match the context synergy pairs are shown in Figures \ref{fig:unseen_only_example} and \ref{fig:seen_example}. These show the retrieved drug after $i$ context examples have been observed by the model. Additionally, we show overall retrieval performance as the number of context examples shown to the model increases in Appendix \ref{appendix:inverse_design}. Figure \ref{fig:ranks_all}. For the weighted strategy, mean rank for seen drugs decreases from $\sim$1,500 to $\sim$400 as context increases. 
Qualitatively, we find that we can retrieve the relevant drug or a similar structure from synergy relationships in multiple cases. This is considerably more effective for drugs observed during training, but performance is also better than random for unknown drugs. This ability to visualize the model's understanding is helpful for explaining what the model predicts from observing a given context. 
Second, it enables retrieving drugs which have a desired set of synergies, which can help inform drug candidate discovery, including patient-specific scenarios. We note that we worked off a broad definition of drug design as discovering new candidate medications. While retrieval is currently a challenging version of this, future work can expand the search space with generative models.

%% file: 5related.tex
\section{Related Work}
\label{appendix:related}

\subsection{Molecular Language Models}
In recent years, advances in machine learning and NLP have been applied to molecule representations. Several efforts \citep{fabian2020molecular, chithrananda2020chemberta, vaucher2021inferring, schwaller2021mapping, megamolbartv2, tysinger2023can} show excellent results training on string representations of molecules \citep{weininger1988smiles, weininger1989smiles, krenn2020self, cheng2023group}. Interest has also grown in multi-modal models \citep{edwards2022translation, zeng2022deep} and multi-encoder models \citep{edwards2021text2mol, vall2021bioassayclr, xu2022protranslator, su2022molecular, liu2022multi, seidl2023enhancing, xu2023protst, zhao2023adversarial} with applications to chemistry and biology. Existing work \citep{edwards2022translation, su2022molecular, xu2023multilingual, christofidellis2023unifying} also builds on this to ``translate'' between these modalities, such as MolT5 \citep{edwards2022translation}, which translates between molecules and language.

\subsection{In-Context Learning}

With the success of models such as GPT-3 \citep{brown2020language} and GPT-4 \citep{openai2023gpt4}, interest has grown in the theoretical properties of in-context learning. \citep{garg2022can}, which we follow in this work, investigates the ability of transformers to learn function classes. \citep{olsson2022context} investigates whether in-context learning is related to specific ``induction heads''. \citep{von2022transformers} shows that transformers do in-context learning by gradient descent. \citep{li2023transformers} frames in-context learning as algorithm learning to investigate generalization on unseen tasks. 

\subsection{Language Models for Chemistry and Knowledge Graph Completion}

Very recently, considerable interest has grown in using language models, particularly GPT-4 \citep{openai2023gpt4}, for uncovering chemical knowledge and molecular discovery \citep{hocky2022natural, white2022large, bran2023chemcrow, boiko2023emergent, white2023assessment, castro2023large}, including work in the few-shot setting \citep{ramos2023bayesian, jablonka2023gpt}. 
CancerGPT \citep{li2023cancergpt}, a related contemporaneous preprint, was recently released which explores a similar few-shot approach to drug-drug synergy prediction. It explores training literature-aware text-based GPT models on drug synergy data. The use of GPT models pretrained on massive textual corpora from the web also makes rigorous evaluation and comparison difficult. We believe our work is complementary, since we largely explore the transformer architecture without language and  we consider in-context learning which they do not. We also consider extensions such as inverse design and context optimization. Due to the recency of \citep{li2023cancergpt}, we leave additional comparisons beyond our real GPT2 baseline to future work. 
Applying language models to knowledge graphs has been investigated in the general \citep{yao2019kg, kim2020multi, youn2022kglm} and scientific domains \citep{nadkarni2021scientific, safavi2022cascader}. They can be considered similar to our tests of BERT language models applied to a drug synergy hypergraph (§~\ref{results:bert}).

\subsection{Drug Synergy Prediction}
As discussed above, there are several approaches \citep{preuer2018deepsynergy, xu2019mr, nyamabo2021ssi, wang2022deepdds, kuru2021matchmaker, sun2020structure,rozemberczki2022chemicalx} which can predict synergy scores given cell line and drug features. There has also been interest in learning representations for these settings \citep{scherer2022distributed}. Recently, work \citep{yang2021graphsynergy, rozemberczki2022moomin, lin2022pisces} has begun to incorporate additional data sources such as drug-protein interactions. This can help improve results, but it often requires creating a subset of the original synergy dataset which can bias results towards the proposed method. \citep{yang2023bliam} extracts additional training data from the literature to improve synergy prediction results, which may relate to our results in Appendix \ref{appendix:scaling}. Research also investigates the application of few-shot \citep{ma2021few} and zero-shot \citep{huang2023zero} machine learning to drug response prediction--we extend this idea to drug synergy prediction. \citep{yang2020stratification} and \citep{kuenzi2020predicting} are related but have different focuses compared to our paper; neither compare against any other synergy baselines or do large-scale evaluation. \citep{yang2020stratification} focuses on a mechanistic understanding of (drug, tumor) activity–a different task. They use this understanding to rank subsystems and predict a limited number of drug combinations to evaluate. \citep{kuenzi2020predicting} does database and experimental testing with small numbers of cell tissues and drugs.

%% file: 6conclusion.tex
\section{Conclusions and Future Work}

As demonstrated by HIV, HCV, and now cancer, combination therapy is a critical option for disease treatment. Yet, difficulties arise in regards to understanding drug-drug interactions and patient-specific genetic differences. To tackle this, we show that encoder-only language models are effective for drug synergy prediction. We then build on these results by proposing SynerGPT, a decoder model with a novel training strategy for in-context learning which can produce strong results for few-shot drug synergy prediction. We additionally show that the model context can be optimized using non-linear black-box approaches, which has exciting implications for the design of a standardized drug synergy testing panel for creating patient-specific synergy datasets. Finally, we explore a novel task of inverse design using desired drug synergy tuples. Performance on this challenging task is low for unknown drugs; nonetheless, it shows promise for future work that may enable personalized drug discovery. %

%% file: 7limitations.tex
\section*{Limitations} \label{limitations}
While we are able to achieve strong performance without additional cellular or drug data, our approach is very much a black box akin to most deep learning methods. To address this, we propose the task of inverse design from drug synergy examples, which allows the visualization of the model's structural understanding as it gains more information. While this is a useful step, we do recognize that further research on mechanistic explainability would be valuable. We hope our contribution on synergy-based inverse design can inspire further work on explainability and that SynerGPT’s predictions can be useful inspiration for clinical researchers. We would also like to note that regardless of using deep learning models, pharmaceutical researchers are in many cases unable to explain the mechanisms of many important drugs on their own (e.g., Modafinil, Metformin, general anesthetics) \citep{stahl2020prescriber, rena2013molecular, brown2011general} --- let alone explain their interactions with each other. These drugs are prescribed to hundreds of millions of patients. Recent studies \citep{lin2019off} suggest that many purported protein drug targets may not be the actual target at all. Important progress with life-saving modern drugs can be made with limited visibility into underlying mechanisms, yet certainly improved mechanistic understanding would be highly useful.

While we show that strong performance is possible without features, future work will still likely want to integrate external database features into drug synergy prediction; however, they will likely need to be integrated in a more thoughtful manner in order to ensure an actual benefit.

It would also likely be interesting for future work to investigate the internal connections language models are learning and what it might mean for understanding the fundamental biology of how cellular pathways interact. It is also worth noting that designing molecules using drug synergy tuples is a somewhat atypical task, so there may exist a wall in terms of the information content inherent in the context. 
While we do analysis by separating model performance into different tissue types in this work (as done in multiple prior studies), we note that for future research it is likely too limiting and simplistic to separate cell lines into tissues types.

%% file: appendix/full_results.tex
\section{Full Results Tables}
\FloatBarrier

\subsection{GraphSynergy Full Results}
\label{appendix:graphsynergy_results}

Full results for the BERT input method and GraphSynergy tests are in Table \ref{tab:full_graphsynergy_transductive_results}. We compare on the specific subset of DrugCombDB \cite{liu2020drugcombdb} which was selected to match Graphsynergy's network data (i.e. selecting the subset of DrugCombDB with drugs/cells that can be matched with external protein-protein interaction, drug-protein association, and cell-protein association networks) and a 7:1:2 train:validation:test split. This data subset also contains useful surface names (the common natural language name of the drug; e.g. dasatinib), which allows us to compare the effect that drug names have on language model synergy prediction performance.

We consider three BERT training variations: the original BERT \cite{devlin2019bert}, SciBERT \cite{beltagy2019scibert}, and BioLinkBERT \cite{yasunaga2022linkbert}. SciBERT was trained on a corpus of scientific documents which would be considerably more focused on drugs than a general corpus. BioLinkBERT is a biomedical BERT model additionally trained using document relation prediction (e.g. citation links). We would like to reiterate the rather remarkable finding that pre-training on scientific literature does not necessarily help the model perform drug synergy prediction any better. Overall, these results indicate that models using external data may not be behaving how we think they are.

\paragraph{ChemicalX Results} We report full results on the subset of DrugCombDB \cite{liu2020drugcombdb} used by ChemicalX \cite{rozemberczki2022chemicalx} in Table \ref{tab:full_transductive_results}. Previous work tested on different subsets of existing datasets (due to filtering for external features).

\begin{table}[h!]%
\centering
\begin{tabular}{ c|c|c|c|c|c|c }

 Input & Model & ROC-AUC & F1 & Precision & Recall & Accuracy \\
\thickhline
  & GraphSynergy & 83.4 & 72.7 & 73.5 & 71.9 & 75.5 \\
 \cline{2-7}
\thickhline
 \multirow{5}{*}{Name} & Unpretrained BERT-base & 80.6 & 
71.0 & 71.7 & 70.3 & 74.0 \\
 \cline{2-7}
  & BioLinkBERT-base & 83.6 & 73.1 & 73.4 & 72.8 & 75.7 \\
 \cline{2-7}
  & SciBERT-base & 83.8 & 73.8 & 73.3 & 74.3 & 75.8 \\
 \cline{2-7}
  & BERT-base & 83.8 & 73.3 & 74.2 & 72.4 & 76.1 \\
  \cline{2-7}
  & BioLinkBERT-large & 84.7 & 73.9 & 74.7 & 73.1 & 76.7 \\
 \cline{2-7}
 \thickhline
 \multirow{4}{*}{\thead{Random\\Token}} & BioLinkBERT-base & 84.1 & 73.7 & 73.6 & 73.8 & 76.2 \\
  \cline{2-7}
 & SciBERT-base & 83.8 & 73.3 & 74.2 & 72.4 & 76.2 \\
 \cline{2-7}
 & BERT-base & 84.0 & 73.4 & 74.1 & 72.7 & 76.1 \\ 
 \cline{2-7}
 & BioLinkBERT-large & 84.1 & 73.8 & 73.4 & 74.2 & 76.1 \\

\end{tabular}
\caption{Performance of BERT models with names and random tokens and GraphSynergy on the custom subset of DrugCombDB \cite{liu2020drugcombdb}. Results are average of 5 runs. Name indicates that the common name of the drug is used as input, while Random Token uses the strategy described in Section \ref{method:encoder}.}
\label{tab:full_graphsynergy_transductive_results}
\end{table}

\begin{table}[h!]%
\centering
\begin{tabular}{ c|c|c|c|c}%

 Model & \thead{KB Info} & \thead{Name Info} & ROC-AUC & PR-AUC \\
\thickhline
 DeepSynergy & $\times$ &  & 84.3 & 70.4 \\
 \hline
 MR-GNN & $\times$ &  & 77.9 & 62.6 \\
 \hline
 SSI-DDI & $\times$ &  & 63.3 & 41.4 \\
 \hline
 DeepDDS & $\times$ & & 87.2 & 77.0 \\
 \hline
 SciBERT (random) & & & 86.9 & 76.3 \\
 \hline
 BioLinkBERT (random) & & & 86.8 & 76.4 \\
 \hline
 BioLinkBERT (name) & & $\times$ & 86.4 & 75.9 \\

\end{tabular}
\caption{Classification results for four selected ChemicalX \cite{rozemberczki2022chemicalx} baselines and two BERT-base models on DrugCombDB \cite{liu2020drugcombdb}. First two BERT models use random token inputs and last model uses drug names as input. Values are average of five runs.}
\label{tab:full_transductive_results}
\end{table}

\FloatBarrier

\subsection{Few-Shot Full Results}
\label{appendix:few_shot_full_results}

\subsubsection{Baseline Descriptions} 
\label{appendix:baseline_descriptions}
DeepSynergy is a popular feedforward model which uses cell line features and drug fingerprints. MR-GNN is a graph convolutional network (GCN) \cite{kipf2016semi} fed into an LSTM \cite{hochreiter1997long} which takes the drug structure into account. SSI-DDI uses a graph attention network (GAT) \cite{velivckovic2017graph} with a final co-attention layer. DeepDDS uses both a GAT and GCN, which are fed into a fully connected feed forward network.

\paragraph{Real GPT-2}
We train a GPT-2 model\footnote{\: ``gpt2'' from HuggingFace.} in the few-shot setting (as opposed to SynerGPT's zero-shot) using random context and the same hyperparameters to mimic SynerGPT's training settings as much as possible. We use names of the drugs obtained from linking to PubChem \cite{kim2023pubchem} as input in the form ``Are drugs [DRUG1] and [DRUG2] synergistic in cell line [CELL]?''. 

\paragraph{SetFit} Furthermore, we test finetuning a few-shot language-model baseline, SetFit \cite{tunstall2022efficient}, on our few-shot data. We follow the original paper in using batch size 16, $R = 20$ text pairs generated for contrastive learning, and 1 epoch. Inputs to the model follow the same format as BERT in Section \ref{method:encoder}. We test using four models. 
\begin{enumerate}
    \item \textbf{SetFit-SBERT}: \textit{paraphrase-multilingual-mpnet-base-v2} from \cite{reimers2019sentence} with names as input. This model was trained to create semantic embeddings via Siamese networks.
    \item \textbf{SetFit-C}: \textit{recobo/chemical-bert-uncased-simcse} from Recobo.ai \footnote{\url{www.recobo.ai}} with names as input. This model was trained using SimCSE on chemistry text.
    \item \textbf{SetFit-S2}: \textit{allenai/specter2} from \cite{singh2022scirepeval} with names as input. This model was trained on multiple scientific classification and regression tasks, such as MeSH descriptors classification.
    \item \textbf{SetFit-SMILES}: \textit{DeepChem/ChemBERTa-77M-MTR} from \cite{ahmad2022chemberta} with SMILES strings as input. This model was pretrained by predicting 200 molecular properties for a molecule given its SMILES string.
\end{enumerate}

\paragraph{Model-Agnostic Meta-Learning} We also consider a meta-learning formulation of our problem setting. We use MAML \cite{finn2017model} to train a DeepDDS model. Since MAML\footnote{We use the implementation from \url{https://github.com/cnguyen10/few_shot_meta_learning}} does few-shot classification using episodes sampled from different learning tasks, we reframe our problem to match this. We consider predicting synergy for each drug to be a task. Then, we sample an episode for training from a random task for each mini-batch. We aggregate rare drugs without enough samples to form an episode into the same task until there are enough samples for an episode. Additionally, since we are dealing with binary classification here, we use $N=2$-way. We sample the ``validation'' portion of each episode from our training set like in SynerGPT. We use the same context bank (and context size) for ``adaptation'' during evaluation. The same learning rate ($1e-3$), batch size (512), and number of steps/epochs as DeepDDS is used. We report few-shot (first-order) and zero-shot (no adaptation) versions. Overall, we find that the MAML training procedure produces poor results, and adaptation produces insignificant performance increases. We attribute this to the episode-based sampling strategy neglecting important information in training. 

\paragraph{Protonets}
As another meta-learning baseline, we consider Protonets \cite{snell2017prototypical}. We use the same meta-learning framework as for MAML. Because we don't have drug task meta-data, we only consider the few-shot setting. %

\paragraph{k-Nearest Neighbors}
We also consider a k-Nearest Neighbors baseline using scikit-learn \cite{scikit-learn} similar to \cite{nadkarni2021scientific}. We construct embeddings for each synergy pair by concatenating (Drug1, Drug2, Cell) embeddings. In the training set, we also include (Drug2, Drug1, Cell). We consider two embedding sources. For the first, kNN-Features, we consider the drug and cell fingerprint features from ChemicalX. For the second, kNN-S2, we use name embeddings from the Specter2 model. We report both zero-shot and few-shot versions. In the few-shot setting, the context bank is added to the training data. We set $k$ equal to the context number (20 and 10 for drugs and cell lines, respectively). We find performance on cell lines to be surprisingly effective, although still less than SynerGPT.

\subsubsection{Interpolate Details}
\label{appendix:interpolate}
In the Unknown cell line setting, we observe that Random has an interesting effect where it performs better after examples (although still worse than Unknown-First (no-ex)), so we consider a fourth strategy: interpolating between Random Unknown-First. 
Essentially, for each data mini-batch in epoch $e$ of $E$ total epochs, we select either the Random strategy with probability $\max(0.25, 1-\frac{e}{E})$ otherwise we use the Unknown-First Strategy. This is analogous to an exploration-exploitation approach where we are pretraining with Random and transitioning to Unknown-First. We use a threshold of 25\% to ensure the benefits of Random are kept until the end of training. 
We find that this interpolation strategy is effective (with p < 0.05, see Table \ref{tab:full_inductive_results}) in dealing with the unknown cell line case.

\begin{table}
\centering

\begin{tabular}{ c|c|c|c||c|c }

 \multicolumn{2}{c}{} & \multicolumn{2}{c}{\underline{Unknown Drug}} & \multicolumn{2}{c}{\underline{Unknown Cell Line}} \\ 
Mode & \multicolumn{1}{c}{Model} & \multicolumn{1}{c}{ROC-AUC} & PR-AUC & ROC-AUC & \multicolumn{1}{c}{PR-AUC} \\
\thickhline
 \multirow{9}{*}{Zero-Shot} & DeepSynergy & 67.5 & 47.7 & 78.6 & 63.6 \\ 
 \cline{2-6}
 & MR-GNN & 65.9 & 44.7 & 76.6 & 61.9 \\
 \cline{2-6}
 & SSI-DDI & 61.8 & 38.9 & 66.6 & 46.7 \\
 \cline{2-6}
 & DeepDDS & 72.1 & 53.2 & 74.5 & 59.8 \\
 \cline{2-6}
 & SciBERT & 67.7 & 47.4 & 79.1 & 64.4 \\
 \cline{2-6}
 & BioLinkBERT & 65.8 & 45.6 & 79.0 & 64.5 \\
 \cline{2-6}
 & MAML-DeepDDS & 68.76 & 50.05 & 71.6 & 54.6 \\
 \cline{2-6}
 & kNN-Features & 65.4 & 45.9 & 82.0 & 70.3 \\
 \cline{2-6}
 & kNN-S2 & 69.2 & 49.0 & 78.8 & 66.0 \\
 \thickhline
 \multirow{15}{*}{Few-Shot} & DeepSynergy & 71.6 & 53.9 & 82.0 & 68.7 \\
 \cline{2-6}
 & MR-GNN & 68.1 & 48.4 & 76.5 & 62.1 \\
 \cline{2-6}
 & SSI-DDI & 62.8 & 40.5 & 66.2 & 45.6 \\
 \cline{2-6}
 & DeepDDS & 75.5 & 57.4 & 74.2 & 60.4 \\
 \cline{2-6}
 & SciBERT & 73.8 & 56.9 & 80.5 & 66.4 \\
 \cline{2-6}
 & BioLinkBERT & 73.0 & 55.6 & 80.6 & 67.4 \\
 \cline{2-6}
 & GPT-2 & 74.2 & 56.8 & 80.3 & 66.6 \\
 \cline{2-6}
 & SetFit-SBERT & 61.4 & 40.7 & 63.6 & 44.0 \\
 \cline{2-6}
 & SetFit-C & 58.9 & 39.6 & 63.8 & 44.8 \\
 \cline{2-6}
 & SetFit-S2 & 58.8 & 39.4 & 63.3 & 44.6 \\
 \cline{2-6}
 & SetFit-SMILES & 63.6 & 43.6 & 64.6 & 44.5 \\
 \cline{2-6}
 & MAML-DeepDDS & 68.79 & 50.00 & 71.4 & 54.6 \\
 \cline{2-6}
 & Protonets-DeepDDS & 54.5 & 31.1 & 57.2 & 34.3 \\
 \cline{2-6}
 & kNN-Features & 66.9 & 47.7 & 82.1 & 70.5 \\
 \cline{2-6}
 & kNN-S2 & 70.0 & 49.9 & 79.0 & 66.2 \\
 \thickhline
 \multirow{9}{*}{SynerGPT} & Features & 73.4 & 55.5 & 70.7 & 52.7 \\
 \cline{2-6}
  & Random (no-ex) & 72.2 & 54.5 & 77.1 & 61.3  \\
 \cline{2-6}
  & Random & 73.7 & 56.8 & 82.3 & 70.2 \\
 \cline{2-6}
  & Graph (no-ex) & 73.2 & 56.1 & 83.3 & 71.7 \\
 \cline{2-6}
  & Graph & 75.5 & 59.6 & 83.2 & 71.5 \\
 \cline{2-6}
  &  Unknown-First (no-ex) & 74.0 & 57.3  & 82.9 & 71.1 \\
 \cline{2-6}
  &  Unknown-First & \textbf{77.7} & \textbf{61.5} & 81.7 & 69.9 \\
 \cline{2-6}
  & Interpolate (no-ex) &  &  & 83.5 & 72.1 \\
 \cline{2-6}
  & Interpolate & &  & \textbf{83.8} & \textbf{72.8}  \\

\end{tabular}

\caption{Few-shot and zero-shot results on ChemicalX DrugCombDB subset with 50 unknown drugs (left) and 20 unknown cell lines (right). Results are the average of 5 runs. no-ex indicates that our trained SynerGPT models were evaluated without any context examples. Features is a SynerGPT model where drug and cell line features are used instead of a randomly-initialized embedding layer. BERT models use random token inputs. Results are the average of 5 runs. The difference between Unknown-First with and without context has p < 0.05 for unknown drugs based on a paired \textit{t}-test. On unknown cell lines using the interpolate strategy, p < 0.05. Similarly, both are statistically significant from the best baseline. %
}
\label{tab:full_inductive_results}
\end{table}

\subsubsection{In-Context Implementation Details}

In the unknown drug setting, to allow for tuples with multiple unknown drugs, we use both a \textbf{[UNKNOWN]} and \textbf{[UNKNOWN2]} token (e.g. a tuple containing two unknown drugs would be $(\textbf{[UNKNOWN]}, \textbf{[UNKNOWN2]}, c)$). 

For the inverse design experiments, in some cases, context examples do not contain the unknown entity $h$ and therefore no \textbf{[UNKNOWN]} tokens, so we use $\overrightarrow{0}$ as a replacement for the ground truth representation when calculating our loss function. We use the same model, splits, and training hyperparameters as in the context optimization setting.

%% file: appendix/optimization.tex
\section{Context Optimization}
\label{appendix:optimization}

\begin{table}%
\centering
\begin{subtable}{0.75\textwidth}
\resizebox{\columnwidth}{!}{
\begin{tabular}{ c|c|c|c|c }

 \multicolumn{1}{c}{} & \multicolumn{2}{c}{\underline{Unknown Drug}} & \multicolumn{2}{c}{\underline{Unknown Cell Line}} \\ 
 \multicolumn{1}{c}{Strategy} & \multicolumn{1}{c}{ROC-AUC} & PR-AUC & ROC-AUC & \multicolumn{1}{c}{PR-AUC} \\
\thickhline
 Typical Unknown-First & 79.2 & 63.8 & 85.2 & 74.9 \\
 \hline
 Best Unknown-First & 80.8 & 66.4 & 85.6 & 75.7 \\
 \hline
 Error Reduction & 75.4 & 59.0 & 84.9 & 74.5 \\
 \hline
 Genetic Algorithm & \textbf{81.5} & \textbf{66.9} & \textbf{86.1} & \textbf{76.5} \\

\end{tabular}}
\end{subtable}
\caption{Test-set performance of different context optimization methods applied to the Unknown-First strategy. Note that the same model parameters are used in all cases and only the input context is changed. Considering Unknown-First as the distribution being sampled from, the genetic algorithm solution has a z-score of 4.02 indicating $p < 0.0001$.}
\label{tab:full_optimization}
\end{table} 

\subsection{Genetic Algorithm}
In the case of the genetic algorithm, each context bank synergy tuple $x \in \mathfrak{D}^c$ is considered as a gene which can be selected by the algorithm. Given $p$ ``unknown'' drugs or cell lines, each has $n$ slots for context examples in its prompt, which makes for $np$ total genes. We also enforce that each $x$ contains the relevant unknown drug $d^h$ or cell line $c^h$. We disallow each context example from being selected multiple times; the reasons for this is two-fold. First, in early experiments we found that if we use the same example for the entire context (e.g. 20 repeats of $x$), then the model performs poorly. This is likely because the model is not trained on duplicate input, so it is trying to make meaningless connections between the same $x$. Second, repeating $x$ in the context provides no new information to the model. Although we enforce this constraint, in practice without it the model will likely do the same thing on its own. 

For the genetic algorithm in context optimization, we use a population of 8 for 50 epochs. We use steady-state parent selection with 4 parents, single-point cross-over, 10\% gene mutation, and elitism. Each example in the context bank is considered a gene and we disallow repeated genes. This results in 351 evaluations on the validation set. 

\subsection{Error Reduction for Context Optimization} \label{appendix:error_reduction}
Using the Unknown-First strategy, we sample a context for some heldout tuple in the validation set. We then calculate the absolute error $\epsilon$ for the heldout tuple. For each context example $x^c$ in the heldout tuple's input context $P^n$, we store $\epsilon$ and the relevant heldout entity, $h$. After some number (for fairness we use the same number of times as the genetic algorithm evaluates ROC-AUC on the validation set--351) of epochs on the validation set, we calculate a mean error $\hat{\epsilon}_{h}(x^c)$ for that context example $x^c$. Finally, for each heldout drug or cell $h$, we select the $n$ context examples $x^c_i$ with the lowest $\hat{\epsilon}_{h}(x^c_i)$. 

As shown in Table \ref{tab:full_optimization}, this strategy produces poor performance. This indicates that simply selecting all of the most individually informative context examples is not useful. Rather, there is a more complex, non-linear interaction between examples which is informative to the model. This is intuitive, because the interaction between cellular pathways in complex and still not well understood. The ability for the context to be optimized by a genetic algorithm but not error reduction indicates that data collection strategies which emphasize diversity may be important to consider for constructing new drug synergy datasets. 

%% file: appendix/GA_change_over_time.tex
\section{Genetic Algorithm Performance}
\label{appendix:GA_change}

\begin{figure}[ht]
\centering
\begin{subfigure}{.8\textwidth}
  \centering
\includegraphics[width=\textwidth]{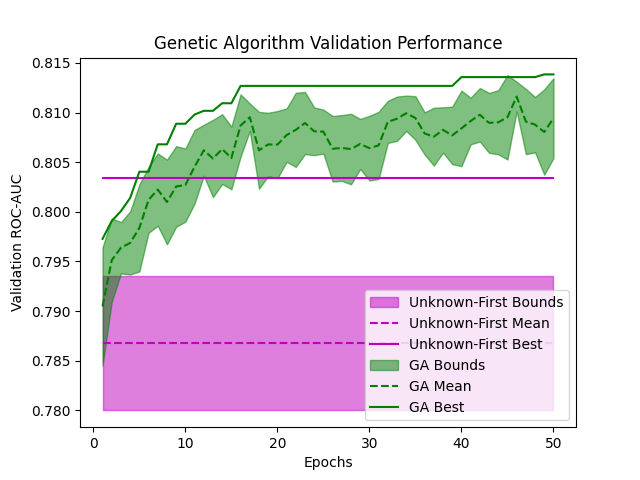}
  \caption{Unknown Drug}
  \label{fig:sub1}
\end{subfigure}

\begin{subfigure}{.8\textwidth}
  \centering
\includegraphics[width=\textwidth]{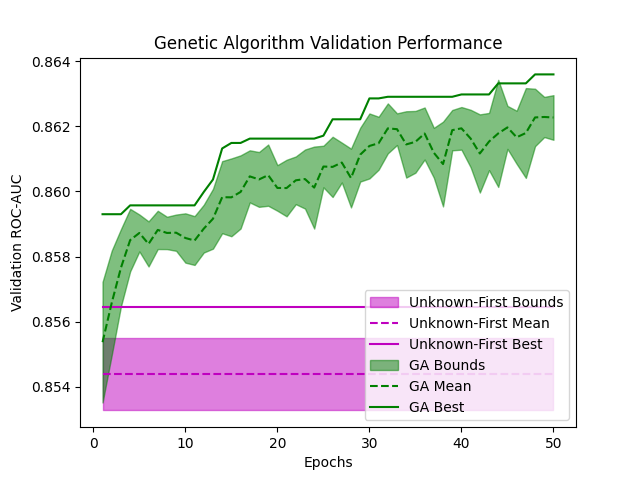}
  \caption{Unknown Cell Line}
  \label{fig:sub2}
\end{subfigure}

\caption{Context optimization performance increase over epochs for the genetic algorithm on both unknown drugs and unknown cell lines. Highlighted areas shows error bounds. Purple regions shows an equivalent number of Unknown-First tests as the genetic algorithm.}
\label{fig:ga_optim}
\end{figure}

%% file: appendix/inverse_design.tex
\section{Inverse Design Additional Results}
\label{appendix:inverse_design}

\begin{figure}[ht]
\centering
\includegraphics[width=\linewidth]{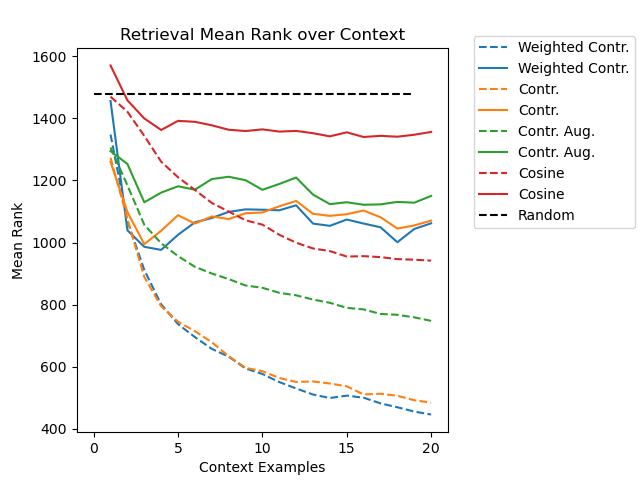}
\caption{This figure shows the mean rank of the retrieved structure as more context synergy tuples are shown to the model. Solid lines show unknown drugs and dotted lines indicate known drugs. Four training variants are shown: Contr. is the contrastive loss described in Equation \ref{clip_loss}, Aug. uses five MegaMolBART representations from augmented SMILES strings for each drug. Cosine uses a simple cosine distance loss. Averaged over 5 runs. See \ref{sec:inverse_design} for details on weighted. }%
\label{fig:ranks_all}
\end{figure}

\begin{figure}[ht]
\centering
\includegraphics[width=\linewidth]{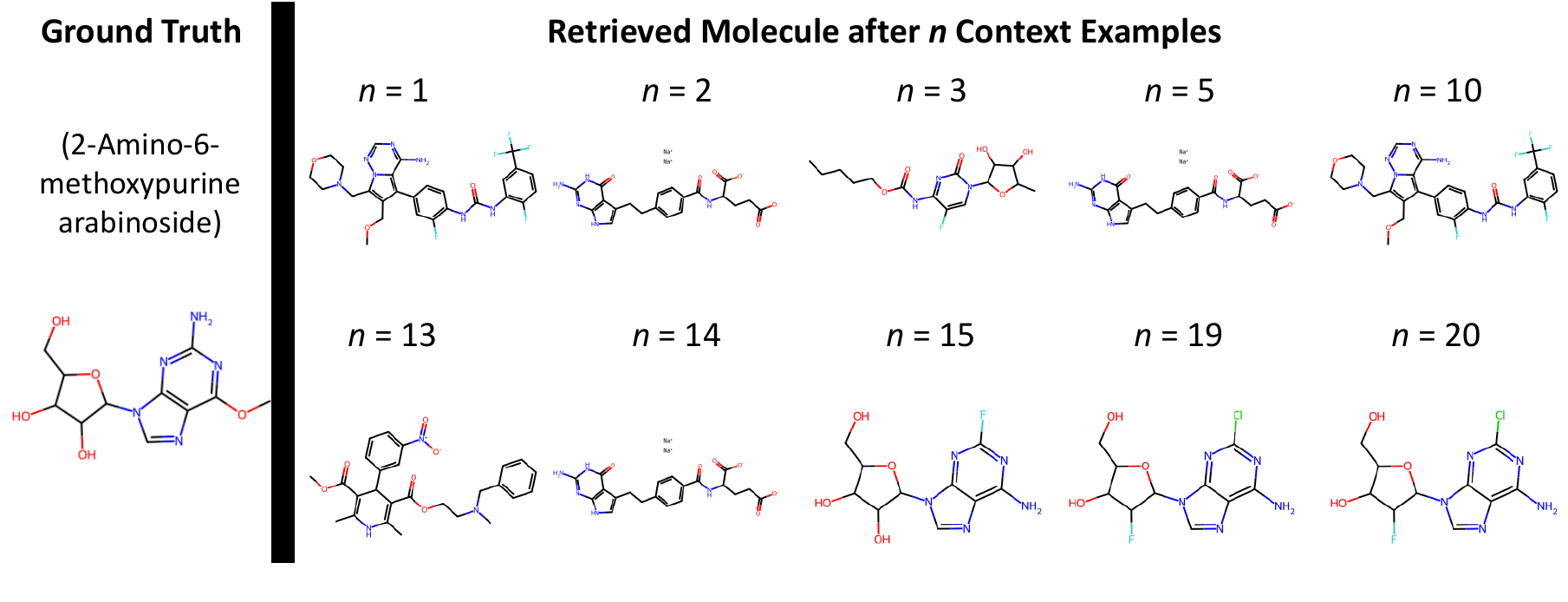}
\caption{This figure shows the model's understanding of a known drug as it observes more context synergy tuples. The ground truth drug is not retrieved, but the model instead identifies two almost structurally identical drugs which would potentially be candidates for interacting with the same drug target. This shows how our method can identify related drugs which behave in similar, synergistic ways. }
\label{fig:seen_example}
\end{figure}

%% file: appendix/tissue_type.tex
\section{Tissue Type Analysis}
\label{appendix:tissue_type}
We further analyze the results of context optimization by separating the results for unknown drugs into their effects on different tissue types. To obtain tissue types, we use the COSMIC \cite{tate2019cosmic} cancer mutation database. Results (Table \ref{tab:drug_tissue_type}) show that performance varies between different tissue types, but that the context optimized by genetic algorithm outperforms default Unknown-First and the model with no context in all cases with exception of pleura. For example, the model excels predicting synergies in ovarian cancers, but results are lower than average in bone and lymphoid cancers. For example, the ROC-AUC of ovarian cancer increases from 77.6 to 81.6\% with examples selected using the default Unknown-First strategy. Our context optimization strategy also shows to be important-- the ROC-AUC further increases significantly to 87.5 using the examples selected by the genetic algorithm. In the unknown cell line case, we see improvements on all cell lines except skin and bone. Interestingly, performance on bone-derived cell lines is low in both settings. 

While we do analysis by separating model performance into different tissue types in this work (as done in multiple prior studies), we note that for future research it is likely too limiting and simplistic to separate cell lines into tissues types. Future studies may look at better bucketing approaches, such as primary cancer-driving mutations. An excellent example are KRAS mutations, which occurr in up to 25\% of human tumors and in many different tissue types (for KRAS: pancreatic, thyroid, colorectal, and lung carcinomas, among others) \cite{kranenburg2005kras}. Further, we note that while our work focuses on few-shot applications to mono-clonal cell lines and tumor biopsies, there is growing evidence that intra-tumor heterogeneity is a driving factor in cancer growth and is also responsible for drug resistance \cite{black2021genetic}. Future work can investigate the effect that this heterogeneity may have on patient-specific drug synergy prediction.

\begin{table}[h!]
\resizebox{\textwidth}{!}{
\centering
\begin{tabular}{ c||c|c||c|c||c|c }

 \multicolumn{1}{c}{} & \multicolumn{2}{c}{\underline{Genetic Algorithm}} & \multicolumn{2}{c}{\underline{No Context}} & \multicolumn{2}{c}{\underline{Typical Unknown-First}} \\
 Tissue Type & ROC-AUC & PR-AUC & ROC-AUC & PR-AUC & ROC-AUC & PR-AUC \\
\thickhline
 skin & \textbf{84.1} & 71.1 & 77.6 & 61.0 & 81.6 & 67.8 \\
 \hline
 ovary & \textbf{87.5} & 80.0 & 81.2 & 73.1 & 86.1 & 77.7 \\
 \hline
 central nervous system & \textbf{82.1} & 66.6 & 76.4 & 54.3 & 79.2 & 61.6 \\
 \hline
 large intestine & \textbf{83.4} & 67.5 & 78.3 & 60.0 & 80.0 & 62.4 \\
 \hline
 pleura & 70.7 & 81.9 & \textbf{82.0} & 90.8 & 68.2 & 81.8 \\
 \hline
 haematopoietic and lymphoid & \textbf{74.1} & 60.5 & 67.9 & 52.7 & 73.6 & 58.2 \\
 \hline 
 lung & \textbf{80.9} & 65.0 & 74.7 & 55.7 & 78.7 & 62.6 \\
 \hline 
 bone & \textbf{69.6} & 65.5 & 69.2 & 65.7 & 68.3 & 65.2 \\
 \hline 
 prostate & \textbf{82.6} & 67.1 & 78.5 & 61.0 & 79.5 & 60.6 \\
 \hline 
 breast & \textbf{84.1} & 68.5 & 75.0 & 54.0 & 80.0 & 64.8 \\
 \hline 
 kidney & \textbf{85.3} & 68.6 & 75.1 & 48.3 & 82.3 & 61.6 \\

\end{tabular}
}
\caption{Unknown drug synergy prediction results separated by tissue type. Results are shown without a context of examples, with the genetic algorithm optimized context, and with the default Unknown-First strategy. }
\label{tab:drug_tissue_type}
\end{table}

\begin{table}[h]
\resizebox{\textwidth}{!}{
\centering
\begin{tabular}{ c||c|c||c|c||c|c }

 \multicolumn{1}{c}{} & \multicolumn{2}{c}{\underline{Genetic Algorithm}} & \multicolumn{2}{c}{\underline{No Context}} & \multicolumn{2}{c}{\underline{Typical Unknown-First}} \\
 Tissue Type & ROC-AUC & PR-AUC & ROC-AUC & PR-AUC & ROC-AUC & PR-AUC \\
\thickhline
 skin & 82.5 & 74.1 & 81.6 & 73.3 & \textbf{83.5} & 75.2 \\
 \hline
 ovary & \textbf{85.1} & 84.8 & 82.0 & 80.9 & 84.0 & 83.4 \\
 \hline
 large intestine & \textbf{88.6} & 81.4 & 87.6 & 79.7 & 87.9 & 80.3 \\
 \hline
 haematopoietic and lymphoid & \textbf{83.8} & 74.0 & 81.8 & 70.2 & 82.3 & 72.2 \\
 \hline
 lung & \textbf{84.2} & 67.2 & 82.9 & 66.9 & 83.2 & 65.3 \\
 \hline 
 bone & 56.2 & 34.6 & 46.3 & 33.9 & \textbf{59.1} & 37.4 \\
 \hline 
 breast & \textbf{89.3} & 80.7 & 88.9 & 79.9 & 88.8 & 79.8 \\
 \hline 
 kidney & \textbf{85.3} & 70.3 & 84.2 & 68.0 & 84.5 & 69.2 \\

\end{tabular}
}
\caption{Unknown cell synergy prediction results separated by tissue type. Base model trained using interpolate strategy and evaluated using Unknown-First.}
\label{tab:cell_tissue_type}
\end{table}

%% file: appendix/scaling.tex
\section{How does training data scale with performance?} 
\label{appendix:scaling}
\begin{figure}[ht]
\centering
\includegraphics[width=\linewidth]{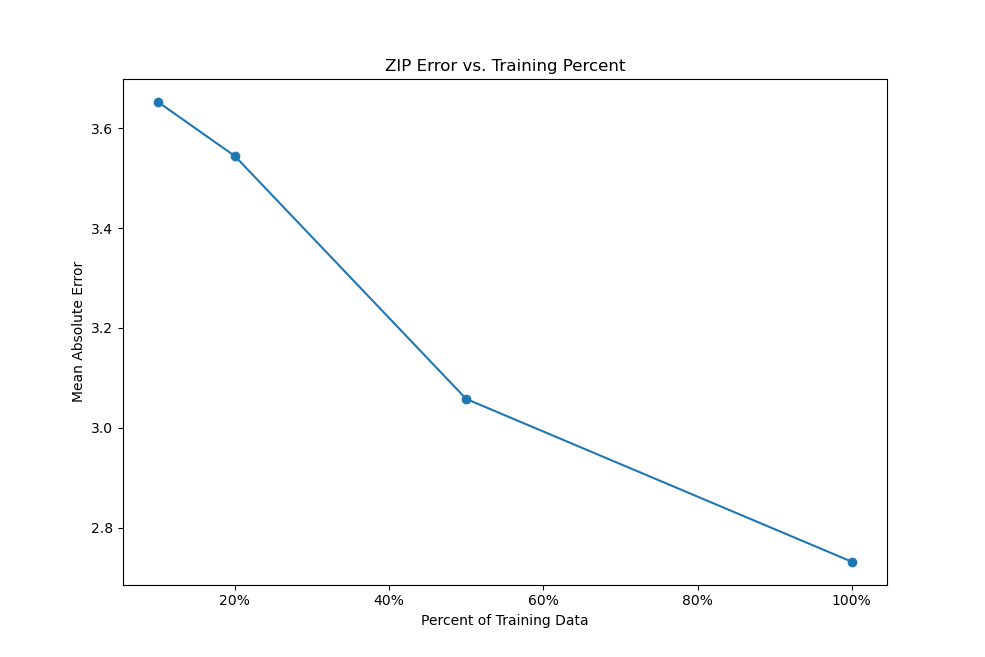}
\caption{Performance versus percent of training data used for BERT with random token input. }
\label{fig:training_percent}
\end{figure}

Figure \ref{fig:training_percent} shows how training data scale affects performance. We consider the version of the DrugComb \cite{zagidullin2019drugcomb} dataset for ZIP score regression recently released in the Therapeutic Data Commons software library \cite{huang2021therapeutics}. It contains 129 drugs, 59 cell lines, and 297,098 synergy tuples. The figure shows BERT model validation performance trained using random token inputs.

%% file: appendix/context_optimization.tex
\section{Additional Diagrams}

\begin{figure*}[ht]
\centering
\includegraphics[width=\linewidth]{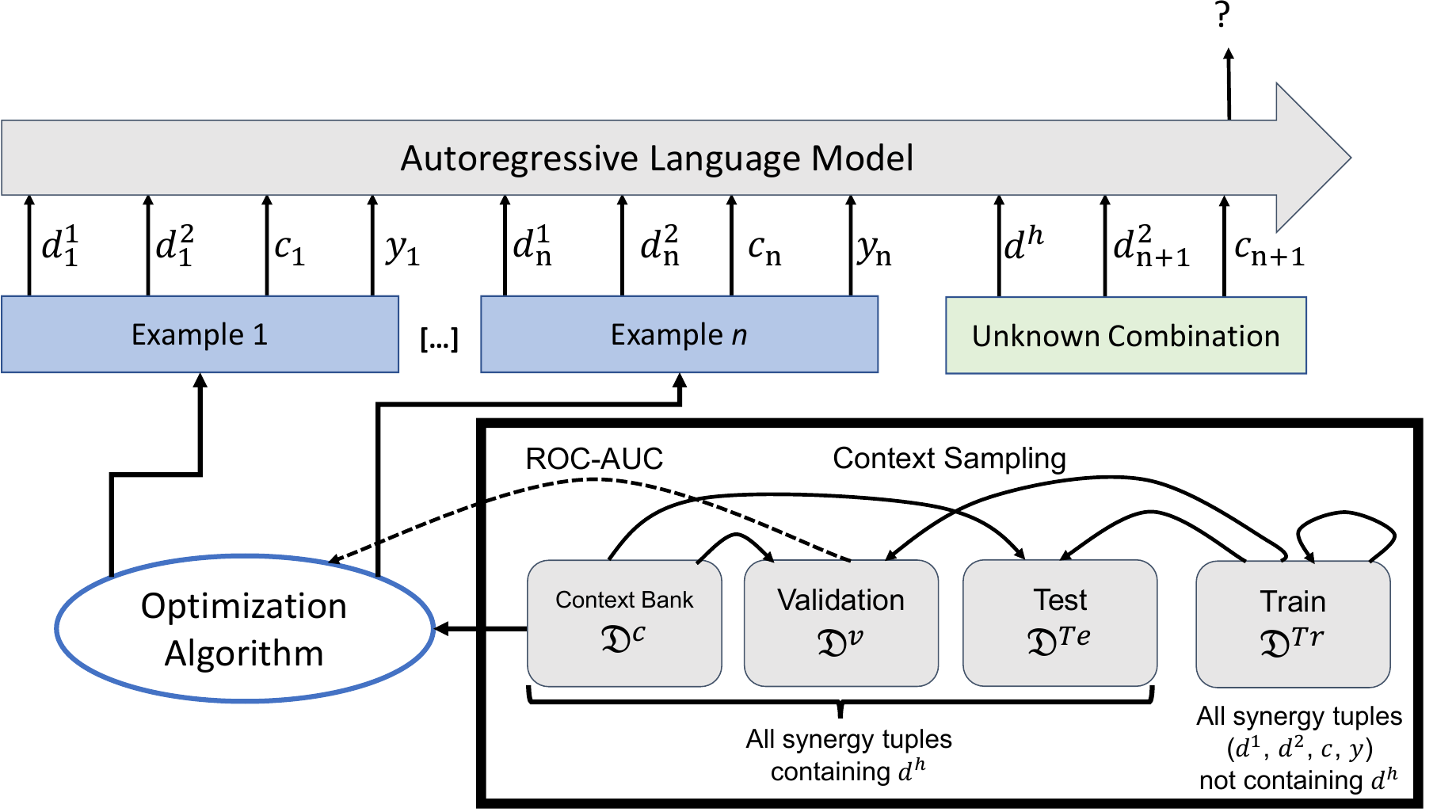}
\caption{The process of optimizing the context (here showing the unknown drug setting). Solid arrows indicate where data can be accessed to build the model prompt. For example, evaluating the model on the test split can use synergy tuples from the train split and context bank. Training the model can only use tuples from the training split. The dotted line indicates that the model's evaluation on that split is used in the optimization algorithm. In our case, this is the ROC-AUC score on the validation set.} %
\label{fig:contextoptim}
\end{figure*}

\begin{figure*}[ht]
\centering
\includegraphics[width=\linewidth]{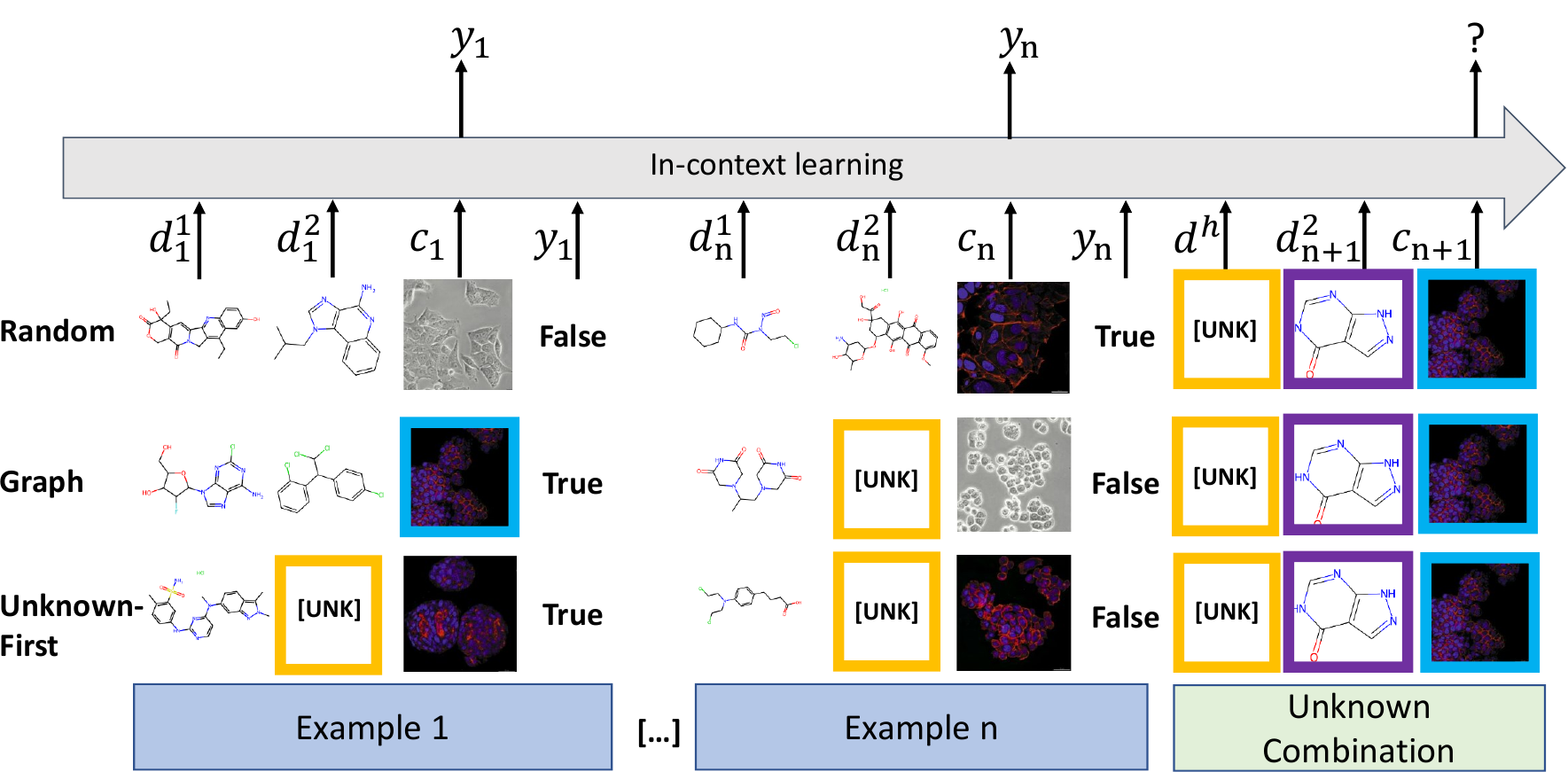}
\caption{Example input of SynerGPT. Figure shows the three in-context learning strategies considered in this work. Colored boxes indicate when the input item is the same. Examples with the same colored box can be thought of as connected nodes in a graph. Example cell line images are from \cite{stankevicius2016extracellular}. Images are for visualization purposes-- in practice a learnable embedding is used as described in \ref{method:incontext}.}
\label{fig:incontext_archi}
\end{figure*}

%% file: appendix/compute.tex
\section{Computing and Implementation Details}

All experiments were done on an internal cluster of GPUs. Each experiment was conducted on a single NVIDIA RTX A6000 with 48 GB VRAM. Notably, multiple experiments can fit on the GPU at one time. Our BERT model experiments done within the ChemicalX framework take roughly 2.5 hours each. For SynerGPT, the Unknown-First and Graph variants took roughly 3 hours to train. Random was compute-bound by sampling, which caused it to take 9 hours to train. The inverse design variants took roughly 6-7 hours to train. We estimate that 80 days of GPU time were used for all experiments. %

BERT-base consists of 108,233,473 parameters and BERT-large is 333,476,865. Unknown drug SynerGPT contains 22,793,473 parameters. Unknown cell version is 18,044,673 parameters. 
BERT-large models (we experimented with BioLinkBERT-large) were unstable to train in many cases. BERT and SynerGPT training used a linear decay learning rate schedule. Unknown drug SynerGPT uses 10,000 steps of warm-up. BERT used 1000 steps of warm-up. Unknown cell SynerGPT used 5\% of training steps as warm-up. The training epoch for unknown drugs is 40 and unknown cell lines is 30. Since we used MegaMolBARTv2 embeddings, we used an output head size of 512 dimensions for retrieval.
SynerGPT is trained using masked context examples selected from the training data– it does not see the unknown drug or cell line during training, which we categorize as zero-shot. Thus, when the SynerGPT model is evaluated on the test set without context examples, it is doing zero-shot prediction for the unknown drug or cell line. When in-context learning is done, it is few-shot. On ChemicalX DrugCombDB, we use a batch size of 512 with random tokens and 256 for names due to VRAM limits. For all random token BERT experiments, we use a high threshold of $k=5,000$ to ensure no common tokens are used.

For the GraphSynergy dataset experiments, BERT-base models use a learning rate of 2e-5. We use 5e-6 for large models, which we find improves training stability. We use a batch size of 32.

Dataset split size varies depending on the seed for evaluating unknown drugs and cell lines. We detail our procedure for building these splits in Section \ref{sec:icl_results} and \ref{sec:context_optim}. This is necessary because we conduct the split based on unknown drugs instead of as percentages. For Table \ref{tab:optimization} experiments, we follow the splitting procedure used in ChemicalX \cite{rozemberczki2022chemicalx}-- this yields on average (145766, 161647) training examples and (44625, 29544) test examples for unknown (drug, cell line), respectively. For the context optimization of unknown drugs and inverse design, the training set size is 145766 and the context, validation, and test set size are each 15208 examples on average.

%% file: appendix/metrics.tex
\section{Evaluation Metrics}

In this section, we detail the binary classification metrics that are used in this paper. Assume we have values for true positives $TP$, true negatives $TN$, false positives $FP$, and false negatives $FN$ where predictions are separated into positives and negatives based on some threshold $t$. 

\begin{itemize}
    \item Accuracy: (TP + TN) / (TP + TN + FP + FN)
    \item Precision: TP / (TP + FP)
    \item Recall: TP / (TP + FN)
    \item TPR: TP / (TP + FN)
    \item FPR: FP / (TN + FP)
    \item TNR: TN / (TN + FP)
    \item ROC-AUC \cite{bradley1997use}: The area under the curve created by plotting TPR against FPR as $t$ is varied.
    \item PR-AUC: Similar to ROC-AUC but the curve is TPR against Precision.
    \item F$_1$: 2TP / (2TP + FP + FN)

\end{itemize}

Given a list of rankings $R$,

$$ Mean Rank = \frac{1}{n}\sum_{i=1}^n R_i $$

%% file: appendix/drugcomb_experiment.tex
\section{DrugComb Language Model Experiments}
\label{appendix:drugcomb}

In this section we consider whether our experiments on BERT hold on the DrugComb dataset \cite{zagidullin2019drugcomb, zheng2021drugcomb}. Here, we use the version used in \cite{rozemberczki2022chemicalx} which contains 4,146 drugs, 288 cell lines, and 659,333 synergy tuples.

\begin{table}[h!]%
\centering
\begin{tabular}{ c|c|c|c|c}%

 Model & \thead{KB Info} & \thead{Name Info} & ROC-AUC & PR-AUC \\
\thickhline
 DeepSynergy & $\times$ &  & 79.7 & 83.2 \\
 \hline
 MR-GNN & $\times$ &  & 68.2 & 74.7 \\
 \hline
 SSI-DDI & $\times$ &  & 55.8 & 62.7 \\
 \hline
 DeepDDS & $\times$ & & 81.7 & 85.8 \\
 \hline
 SciBERT (random) & & & 82.0 & 86.1 \\
 \hline
 BioLinkBERT (random) & & & 82.5 & 86.6 \\

\end{tabular}
\caption{Classification results for four selected ChemicalX \cite{rozemberczki2022chemicalx} baselines and two BERT-base models on DrugComb \cite{zagidullin2019drugcomb, zheng2021drugcomb}. BERT models use random token inputs. Values are average of five runs.}
\label{tab:drugcomb_full_transductive_results}
\end{table}

%% file: main.bbl
\begin{thebibliography}{105}
\providecommand{\natexlab}[1]{#1}
\providecommand{\url}[1]{\texttt{#1}}
\expandafter\ifx\csname urlstyle\endcsname\relax
  \providecommand{\doi}[1]{doi: #1}\else
  \providecommand{\doi}{doi: \begingroup \urlstyle{rm}\Url}\fi

\bibitem[Ahmad et~al.(2022)Ahmad, Simon, Chithrananda, Grand, and Ramsundar]{ahmad2022chemberta}
Walid Ahmad, Elana Simon, Seyone Chithrananda, Gabriel Grand, and Bharath Ramsundar.
\newblock Chemberta-2: Towards chemical foundation models.
\newblock \emph{arXiv preprint arXiv:2209.01712}, 2022.

\bibitem[Beltagy et~al.(2019)Beltagy, Lo, and Cohan]{beltagy2019scibert}
Iz~Beltagy, Kyle Lo, and Arman Cohan.
\newblock Scibert: A pretrained language model for scientific text.
\newblock \emph{arXiv preprint arXiv:1903.10676}, 2019.

\bibitem[Bertsimas et~al.(2015)Bertsimas, King, and Mazumder]{bertsimas2016best}
Dimitris Bertsimas, Angela King, and Rahul Mazumder.
\newblock Best subset selection via a modern optimization lens.
\newblock \emph{arXiv: Methodology}, 2015.

\bibitem[Black \& McGranahan(2021)Black and McGranahan]{black2021genetic}
James~RM Black and Nicholas McGranahan.
\newblock Genetic and non-genetic clonal diversity in cancer evolution.
\newblock \emph{Nature Reviews Cancer}, 21\penalty0 (6):\penalty0 379--392, 2021.

\bibitem[Boiko et~al.(2023)Boiko, MacKnight, and Gomes]{boiko2023emergent}
Daniil~A Boiko, Robert MacKnight, and Gabe Gomes.
\newblock Emergent autonomous scientific research capabilities of large language models.
\newblock \emph{arXiv preprint arXiv:2304.05332}, 2023.

\bibitem[Bradley(1997)]{bradley1997use}
Andrew~P Bradley.
\newblock The use of the area under the roc curve in the evaluation of machine learning algorithms.
\newblock \emph{Pattern recognition}, 30\penalty0 (7):\penalty0 1145--1159, 1997.

\bibitem[Bran et~al.(2023)Bran, Cox, White, and Schwaller]{bran2023chemcrow}
Andres~M Bran, Sam Cox, Andrew~D White, and Philippe Schwaller.
\newblock Chemcrow: Augmenting large-language models with chemistry tools.
\newblock \emph{arXiv preprint arXiv:2304.05376}, 2023.

\bibitem[Brown et~al.(2011)Brown, Purdon, and Van~Dort]{brown2011general}
Emery~N Brown, Patrick~L Purdon, and Christa~J Van~Dort.
\newblock General anesthesia and altered states of arousal: a systems neuroscience analysis.
\newblock \emph{Annual review of neuroscience}, 34:\penalty0 601--628, 2011.

\bibitem[Brown et~al.(2020)Brown, Mann, Ryder, Subbiah, Kaplan, Dhariwal, Neelakantan, Shyam, Sastry, Askell, et~al.]{brown2020language}
Tom Brown, Benjamin Mann, Nick Ryder, Melanie Subbiah, Jared~D Kaplan, Prafulla Dhariwal, Arvind Neelakantan, Pranav Shyam, Girish Sastry, Amanda Askell, et~al.
\newblock Language models are few-shot learners.
\newblock \emph{Advances in neural information processing systems}, 33:\penalty0 1877--1901, 2020.

\bibitem[Castro~Nascimento \& Pimentel(2023)Castro~Nascimento and Pimentel]{castro2023large}
Cayque~Monteiro Castro~Nascimento and Andr{\'e}~Silva Pimentel.
\newblock Do large language models understand chemistry? a conversation with chatgpt.
\newblock \emph{Journal of Chemical Information and Modeling}, 63\penalty0 (6):\penalty0 1649--1655, 2023.

\bibitem[Cheng et~al.(2023)Cheng, Cai, Miret, Malkomes, Phielipp, and Aspuru-Guzik]{cheng2023group}
Austin~H Cheng, Andy Cai, Santiago Miret, Gustavo Malkomes, Mariano Phielipp, and Al{\'a}n Aspuru-Guzik.
\newblock Group selfies: a robust fragment-based molecular string representation.
\newblock \emph{Digital Discovery}, 2023.

\bibitem[Chithrananda et~al.(2020)Chithrananda, Grand, and Ramsundar]{chithrananda2020chemberta}
Seyone Chithrananda, Gabe Grand, and Bharath Ramsundar.
\newblock Chemberta: Large-scale self-supervised pretraining for molecular property prediction.
\newblock \emph{arXiv preprint arXiv:2010.09885}, 2020.

\bibitem[Christofidellis et~al.(2023)Christofidellis, Giannone, Born, Winther, Laino, and Manica]{christofidellis2023unifying}
Dimitrios Christofidellis, Giorgio Giannone, Jannis Born, Ole Winther, Teodoro Laino, and Matteo Manica.
\newblock Unifying molecular and textual representations via multi-task language modelling.
\newblock \emph{arXiv preprint arXiv:2301.12586}, 2023.

\bibitem[Devlin et~al.(2019)Devlin, Chang, Lee, and Toutanova]{devlin2019bert}
Jacob Devlin, Ming-Wei Chang, Kenton Lee, and Kristina Toutanova.
\newblock Bert: Pre-training of deep bidirectional transformers for language understanding.
\newblock In \emph{Proceedings of the 2019 Conference of the North American Chapter of the Association for Computational Linguistics: Human Language Technologies, Volume 1 (Long and Short Papers)}, pp.\  4171--4186, 2019.

\bibitem[Dong et~al.(2022)Dong, Li, Dai, Zheng, Wu, Chang, Sun, Xu, and Sui]{dong2022survey}
Qingxiu Dong, Lei Li, Damai Dai, Ce~Zheng, Zhiyong Wu, Baobao Chang, Xu~Sun, Jingjing Xu, and Zhifang Sui.
\newblock A survey for in-context learning.
\newblock \emph{arXiv preprint arXiv:2301.00234}, 2022.

\bibitem[Dubuisson(2007)]{dubuisson2007hepatitis}
Jean Dubuisson.
\newblock Hepatitis c virus proteins.
\newblock \emph{World journal of gastroenterology: WJG}, 13\penalty0 (17):\penalty0 2406, 2007.

\bibitem[Edwards et~al.(2021)Edwards, Zhai, and Ji]{edwards2021text2mol}
Carl Edwards, ChengXiang Zhai, and Heng Ji.
\newblock {T}ext2{M}ol: Cross-modal molecule retrieval with natural language queries.
\newblock In \emph{Proceedings of the 2021 Conference on Empirical Methods in Natural Language Processing}, pp.\  595--607, Online and Punta Cana, Dominican Republic, November 2021. Association for Computational Linguistics.
\newblock URL \url{https://aclanthology.org/2021.emnlp-main.47}.

\bibitem[Edwards et~al.(2022)Edwards, Lai, Ros, Honke, Cho, and Ji]{edwards2022translation}
Carl Edwards, Tuan Lai, Kevin Ros, Garrett Honke, Kyunghyun Cho, and Heng Ji.
\newblock Translation between molecules and natural language.
\newblock In \emph{Proceedings of the 2022 Conference on Empirical Methods in Natural Language Processing}, pp.\  375--413, Abu Dhabi, United Arab Emirates, December 2022. Association for Computational Linguistics.
\newblock URL \url{https://aclanthology.org/2022.emnlp-main.26}.

\bibitem[Fabian et~al.(2020)Fabian, Edlich, Gaspar, Segler, Meyers, Fiscato, and Ahmed]{fabian2020molecular}
Benedek Fabian, Thomas Edlich, H{\'e}l{\'e}na Gaspar, Marwin Segler, Joshua Meyers, Marco Fiscato, and Mohamed Ahmed.
\newblock Molecular representation learning with language models and domain-relevant auxiliary tasks.
\newblock \emph{arXiv preprint arXiv:2011.13230}, 2020.

\bibitem[Finn et~al.(2017)Finn, Abbeel, and Levine]{finn2017model}
Chelsea Finn, Pieter Abbeel, and Sergey Levine.
\newblock Model-agnostic meta-learning for fast adaptation of deep networks.
\newblock In \emph{International conference on machine learning}, pp.\  1126--1135. PMLR, 2017.

\bibitem[Frankel \& Young(1998)Frankel and Young]{frankel1998hiv}
Alan~D Frankel and John~AT Young.
\newblock Hiv-1: fifteen proteins and an rna.
\newblock \emph{Annual review of biochemistry}, 67\penalty0 (1):\penalty0 1--25, 1998.

\bibitem[Gad(2021)]{gad2021pygad}
Ahmed~Fawzy Gad.
\newblock Pygad: An intuitive genetic algorithm python library, 2021.

\bibitem[Garg et~al.(2022)Garg, Tsipras, Liang, and Valiant]{garg2022can}
Shivam Garg, Dimitris Tsipras, Percy~S Liang, and Gregory Valiant.
\newblock What can transformers learn in-context? a case study of simple function classes.
\newblock \emph{Advances in Neural Information Processing Systems}, 35:\penalty0 30583--30598, 2022.

\bibitem[Hochreiter \& Schmidhuber(1997)Hochreiter and Schmidhuber]{hochreiter1997long}
Sepp Hochreiter and J{\"u}rgen Schmidhuber.
\newblock Long short-term memory.
\newblock \emph{Neural computation}, 9\penalty0 (8):\penalty0 1735--1780, 1997.

\bibitem[Hocky \& White(2022)Hocky and White]{hocky2022natural}
Glen~M Hocky and Andrew~D White.
\newblock Natural language processing models that automate programming will transform chemistry research and teaching.
\newblock \emph{Digital discovery}, 1\penalty0 (2):\penalty0 79--83, 2022.

\bibitem[Huang et~al.(2021)Huang, Fu, Gao, Zhao, Roohani, Leskovec, Coley, Xiao, Sun, and Zitnik]{huang2021therapeutics}
Kexin Huang, Tianfan Fu, Wenhao Gao, Yue Zhao, Yusuf Roohani, Jure Leskovec, Connor~W Coley, Cao Xiao, Jimeng Sun, and Marinka Zitnik.
\newblock Therapeutics data commons: Machine learning datasets and tasks for drug discovery and development.
\newblock \emph{arXiv preprint arXiv:2102.09548}, 2021.

\bibitem[Huang et~al.(2023)Huang, Chandak, Wang, Havaldar, Vaid, Leskovec, Nadkarni, Glicksberg, Gehlenborg, and Zitnik]{huang2023zero}
Kexin Huang, Payal Chandak, Qianwen Wang, Shreyas Havaldar, Akhil Vaid, Jure Leskovec, Girish Nadkarni, Benjamin~S Glicksberg, Nils Gehlenborg, and Marinka Zitnik.
\newblock Zero-shot prediction of therapeutic use with geometric deep learning and clinician centered design.
\newblock \emph{medRxiv}, pp.\  2023--03, 2023.

\bibitem[Jablonka et~al.(2023)Jablonka, Schwaller, Ortega-Guerrero, and Smit]{jablonka2023gpt}
Kevin~Maik Jablonka, Philippe Schwaller, Andres Ortega-Guerrero, and Berend Smit.
\newblock Is gpt-3 all you need for low-data discovery in chemistry?
\newblock \emph{ChemRxiv preprint}, 2023.

\bibitem[Jin et~al.(2020)Jin, Barzilay, and Jaakkola]{jin2020hierarchical}
Wengong Jin, Regina Barzilay, and Tommi Jaakkola.
\newblock Hierarchical generation of molecular graphs using structural motifs.
\newblock In \emph{International conference on machine learning}, pp.\  4839--4848. PMLR, 2020.

\bibitem[Juurlink et~al.(2004)Juurlink, Mamdani, Lee, Kopp, Austin, Laupacis, and Redelmeier]{juurlink2004rates}
David~N Juurlink, Muhammad~M Mamdani, Douglas~S Lee, Alexander Kopp, Peter~C Austin, Andreas Laupacis, and Donald~A Redelmeier.
\newblock Rates of hyperkalemia after publication of the randomized aldactone evaluation study.
\newblock \emph{New England Journal of Medicine}, 351\penalty0 (6):\penalty0 543--551, 2004.

\bibitem[Kim et~al.(2020)Kim, Hong, Ko, and Seo]{kim2020multi}
Bosung Kim, Taesuk Hong, Youngjoong Ko, and Jungyun Seo.
\newblock Multi-task learning for knowledge graph completion with pre-trained language models.
\newblock In \emph{Proceedings of the 28th International Conference on Computational Linguistics}, pp.\  1737--1743, 2020.

\bibitem[Kim et~al.(2023)Kim, Chen, Cheng, Gindulyte, He, He, Li, Shoemaker, Thiessen, Yu, et~al.]{kim2023pubchem}
Sunghwan Kim, Jie Chen, Tiejun Cheng, Asta Gindulyte, Jia He, Siqian He, Qingliang Li, Benjamin~A Shoemaker, Paul~A Thiessen, Bo~Yu, et~al.
\newblock Pubchem 2023 update.
\newblock \emph{Nucleic Acids Research}, 51\penalty0 (D1):\penalty0 D1373--D1380, 2023.

\bibitem[Kipf \& Welling(2016)Kipf and Welling]{kipf2016semi}
Thomas~N Kipf and Max Welling.
\newblock Semi-supervised classification with graph convolutional networks.
\newblock \emph{arXiv preprint arXiv:1609.02907}, 2016.

\bibitem[Kranenburg(2005)]{kranenburg2005kras}
Onno Kranenburg.
\newblock The kras oncogene: past, present, and future.
\newblock \emph{Biochimica et biophysica acta}, 1756\penalty0 (2):\penalty0 81--82, 2005.

\bibitem[Krenn et~al.(2020)Krenn, H{\"a}se, Nigam, Friederich, and Aspuru-Guzik]{krenn2020self}
Mario Krenn, Florian H{\"a}se, AkshatKumar Nigam, Pascal Friederich, and Alan Aspuru-Guzik.
\newblock Self-referencing embedded strings (selfies): A 100\% robust molecular string representation.
\newblock \emph{Machine Learning: Science and Technology}, 1\penalty0 (4):\penalty0 045024, 2020.

\bibitem[Krishna et~al.(2021)Krishna, Bigham, and Lipton]{krishna2021does}
Kundan Krishna, Jeffrey Bigham, and Zachary~C Lipton.
\newblock Does pretraining for summarization require knowledge transfer?
\newblock \emph{arXiv preprint arXiv:2109.04953}, 2021.

\bibitem[Kuenzi et~al.(2020)Kuenzi, Park, Fong, Sanchez, Lee, Kreisberg, Ma, and Ideker]{kuenzi2020predicting}
Brent~M Kuenzi, Jisoo Park, Samson~H Fong, Kyle~S Sanchez, John Lee, Jason~F Kreisberg, Jianzhu Ma, and Trey Ideker.
\newblock Predicting drug response and synergy using a deep learning model of human cancer cells.
\newblock \emph{Cancer cell}, 38\penalty0 (5):\penalty0 672--684, 2020.

\bibitem[Kuru et~al.(2021)Kuru, Tastan, and Cicek]{kuru2021matchmaker}
Halil~Ibrahim Kuru, Oznur Tastan, and A~Ercument Cicek.
\newblock Matchmaker: a deep learning framework for drug synergy prediction.
\newblock \emph{IEEE/ACM Transactions on Computational Biology and Bioinformatics}, 19\penalty0 (4):\penalty0 2334--2344, 2021.

\bibitem[Li et~al.(2023{\natexlab{a}})Li, Shetty, Kamath, Jaiswal, Jiang, Ding, and Kim]{li2023cancergpt}
Tianhao Li, Sandesh Shetty, Advaith Kamath, Ajay Jaiswal, Xianqian Jiang, Ying Ding, and Yejin Kim.
\newblock Cancergpt: Few-shot drug pair synergy prediction using large pre-trained language models.
\newblock \emph{arXiv preprint arXiv:2304.10946}, 2023{\natexlab{a}}.

\bibitem[Li et~al.(2023{\natexlab{b}})Li, Ildiz, Papailiopoulos, and Oymak]{li2023transformers}
Yingcong Li, M.~Emrullah Ildiz, Dimitris Papailiopoulos, and Samet Oymak.
\newblock Transformers as algorithms: Generalization and stability in in-context learning, 2023{\natexlab{b}}.

\bibitem[Liang \& Ghany(2013)Liang and Ghany]{liang2013current}
T~Jake Liang and Marc~G Ghany.
\newblock Current and future therapies for hepatitis c virus infection.
\newblock \emph{New England Journal of Medicine}, 368\penalty0 (20):\penalty0 1907--1917, 2013.

\bibitem[Lin et~al.(2019)Lin, Giuliano, Palladino, John, Abramowicz, Yuan, Sausville, Lukow, Liu, Chait, et~al.]{lin2019off}
Ann Lin, Christopher~J Giuliano, Ann Palladino, Kristen~M John, Connor Abramowicz, Monet~Lou Yuan, Erin~L Sausville, Devon~A Lukow, Luwei Liu, Alexander~R Chait, et~al.
\newblock Off-target toxicity is a common mechanism of action of cancer drugs undergoing clinical trials.
\newblock \emph{Science translational medicine}, 11\penalty0 (509):\penalty0 eaaw8412, 2019.

\bibitem[Lin et~al.(2022)Lin, Xu, Woicik, Ma, and Wang]{lin2022pisces}
Jiacheng Lin, Hanwen Xu, Addie Woicik, Jianzhu Ma, and Sheng Wang.
\newblock Pisces: A combo-wise contrastive learning approach to synergistic drug combination prediction.
\newblock \emph{bioRxiv}, pp.\  2022--11, 2022.

\bibitem[Liu et~al.(2020)Liu, Zhang, Zou, Wang, Deng, and Deng]{liu2020drugcombdb}
Hui Liu, Wenhao Zhang, Bo~Zou, Jinxian Wang, Yuanyuan Deng, and Lei Deng.
\newblock Drugcombdb: a comprehensive database of drug combinations toward the discovery of combinatorial therapy.
\newblock \emph{Nucleic acids research}, 48\penalty0 (D1):\penalty0 D871--D881, 2020.

\bibitem[Liu et~al.(2022)Liu, Nie, Wang, Lu, Qiao, Liu, Tang, Xiao, and Anandkumar]{liu2022multi}
Shengchao Liu, Weili Nie, Chengpeng Wang, Jiarui Lu, Zhuoran Qiao, Ling Liu, Jian Tang, Chaowei Xiao, and Anima Anandkumar.
\newblock Multi-modal molecule structure-text model for text-based retrieval and editing.
\newblock \emph{arXiv preprint arXiv:2212.10789}, 2022.

\bibitem[Ma et~al.(2021)Ma, Fong, Luo, Bakkenist, Shen, Mourragui, Wessels, Hafner, Sharan, Peng, et~al.]{ma2021few}
Jianzhu Ma, Samson~H Fong, Yunan Luo, Christopher~J Bakkenist, John~Paul Shen, Soufiane Mourragui, Lodewyk~FA Wessels, Marc Hafner, Roded Sharan, Jian Peng, et~al.
\newblock Few-shot learning creates predictive models of drug response that translate from high-throughput screens to individual patients.
\newblock \emph{Nature Cancer}, 2\penalty0 (2):\penalty0 233--244, 2021.

\bibitem[Mialon et~al.(2023)Mialon, Dess{\`\i}, Lomeli, Nalmpantis, Pasunuru, Raileanu, Rozi{\`e}re, Schick, Dwivedi-Yu, Celikyilmaz, et~al.]{mialon2023augmented}
Gr{\'e}goire Mialon, Roberto Dess{\`\i}, Maria Lomeli, Christoforos Nalmpantis, Ram Pasunuru, Roberta Raileanu, Baptiste Rozi{\`e}re, Timo Schick, Jane Dwivedi-Yu, Asli Celikyilmaz, et~al.
\newblock Augmented language models: a survey.
\newblock \emph{arXiv preprint arXiv:2302.07842}, 2023.

\bibitem[Miller(2002)]{miller2002subset}
Alan Miller.
\newblock \emph{Subset selection in regression}.
\newblock CRC Press, 2002.

\bibitem[Mitchell et~al.(2007)]{mitchell2007machine}
Tom~Michael Mitchell et~al.
\newblock \emph{Machine learning}, volume~1.
\newblock McGraw-hill New York, 2007.

\bibitem[Mokhtari et~al.(2017)Mokhtari, Homayouni, Baluch, Morgatskaya, Kumar, Das, and Yeger]{mokhtari2017combination}
Reza~Bayat Mokhtari, Tina~S Homayouni, Narges Baluch, Evgeniya Morgatskaya, Sushil Kumar, Bikul Das, and Herman Yeger.
\newblock Combination therapy in combating cancer.
\newblock \emph{Oncotarget}, 8\penalty0 (23):\penalty0 38022, 2017.

\bibitem[Moore \& Chaisson(1999)Moore and Chaisson]{moore1999natural}
Richard~D Moore and Richard~E Chaisson.
\newblock Natural history of hiv infection in the\_era of combination antiretroviral therapy.
\newblock \emph{Aids}, 13\penalty0 (14):\penalty0 1933--1942, 1999.

\bibitem[Mroz \& Rocco(2017)Mroz and Rocco]{mroz2017challenges}
Edmund~A Mroz and James~W Rocco.
\newblock The challenges of tumor genetic diversity.
\newblock \emph{Cancer}, 123\penalty0 (6):\penalty0 917--927, 2017.

\bibitem[Nadkarni et~al.(2021)Nadkarni, Wadden, Beltagy, Smith, Hajishirzi, and Hope]{nadkarni2021scientific}
Rahul Nadkarni, David Wadden, Iz~Beltagy, Noah Smith, Hannaneh Hajishirzi, and Tom Hope.
\newblock Scientific language models for biomedical knowledge base completion: An empirical study.
\newblock In \emph{3rd Conference on Automated Knowledge Base Construction}, 2021.

\bibitem[{NVIDIA Corporation}(2022)]{megamolbartv2}
{NVIDIA Corporation}.
\newblock Megamolbart v0.2, 2022.
\newblock URL \url{https://catalog.ngc.nvidia.com/orgs/nvidia/teams/clara/models/megamolbart_0_2}.

\bibitem[Nyamabo et~al.(2021)Nyamabo, Yu, and Shi]{nyamabo2021ssi}
Arnold~K Nyamabo, Hui Yu, and Jian-Yu Shi.
\newblock Ssi--ddi: substructure--substructure interactions for drug--drug interaction prediction.
\newblock \emph{Briefings in Bioinformatics}, 22\penalty0 (6):\penalty0 bbab133, 2021.

\bibitem[O'Donnell \& Dolan(2009)O'Donnell and Dolan]{o2009cancer}
Peter~H O'Donnell and M~Eileen Dolan.
\newblock Cancer pharmacoethnicity: Ethnic differences in susceptibility to the effects of chemotherapycancer pharmacoethnicity.
\newblock \emph{Clinical Cancer Research}, 15\penalty0 (15):\penalty0 4806--4814, 2009.

\bibitem[Olsson et~al.(2022)Olsson, Elhage, Nanda, Joseph, DasSarma, Henighan, Mann, Askell, Bai, Chen, et~al.]{olsson2022context}
Catherine Olsson, Nelson Elhage, Neel Nanda, Nicholas Joseph, Nova DasSarma, Tom Henighan, Ben Mann, Amanda Askell, Yuntao Bai, Anna Chen, et~al.
\newblock In-context learning and induction heads.
\newblock \emph{arXiv preprint arXiv:2209.11895}, 2022.

\bibitem[OpenAI(2023)]{openai2023gpt4}
OpenAI.
\newblock Gpt-4 technical report, 2023.

\bibitem[Pedregosa et~al.(2011)Pedregosa, Varoquaux, Gramfort, Michel, Thirion, Grisel, Blondel, Prettenhofer, Weiss, Dubourg, Vanderplas, Passos, Cournapeau, Brucher, Perrot, and Duchesnay]{scikit-learn}
F.~Pedregosa, G.~Varoquaux, A.~Gramfort, V.~Michel, B.~Thirion, O.~Grisel, M.~Blondel, P.~Prettenhofer, R.~Weiss, V.~Dubourg, J.~Vanderplas, A.~Passos, D.~Cournapeau, M.~Brucher, M.~Perrot, and E.~Duchesnay.
\newblock Scikit-learn: Machine learning in {P}ython.
\newblock \emph{Journal of Machine Learning Research}, 12:\penalty0 2825--2830, 2011.

\bibitem[Preuer et~al.(2018)Preuer, Lewis, Hochreiter, Bender, Bulusu, and Klambauer]{preuer2018deepsynergy}
Kristina Preuer, Richard~PI Lewis, Sepp Hochreiter, Andreas Bender, Krishna~C Bulusu, and G{\"u}nter Klambauer.
\newblock Deepsynergy: predicting anti-cancer drug synergy with deep learning.
\newblock \emph{Bioinformatics}, 34\penalty0 (9):\penalty0 1538--1546, 2018.

\bibitem[Radford et~al.(2019)Radford, Wu, Child, Luan, Amodei, Sutskever, et~al.]{radford2019language}
Alec Radford, Jeffrey Wu, Rewon Child, David Luan, Dario Amodei, Ilya Sutskever, et~al.
\newblock Language models are unsupervised multitask learners.
\newblock \emph{OpenAI blog}, 1\penalty0 (8):\penalty0 9, 2019.

\bibitem[Radford et~al.(2021)Radford, Kim, Hallacy, Ramesh, Goh, Agarwal, Sastry, Askell, Mishkin, Clark, Krueger, and Sutskever]{radford2021learning}
Alec Radford, Jong~Wook Kim, Chris Hallacy, Aditya Ramesh, Gabriel Goh, Sandhini Agarwal, Girish Sastry, Amanda Askell, Pamela Mishkin, Jack Clark, Gretchen Krueger, and Ilya Sutskever.
\newblock Learning transferable visual models from natural language supervision.
\newblock In Marina Meila and Tong Zhang (eds.), \emph{Proceedings of the 38th International Conference on Machine Learning, {ICML} 2021, 18-24 July 2021, Virtual Event}, volume 139 of \emph{Proceedings of Machine Learning Research}, pp.\  8748--8763. {PMLR}, 2021.
\newblock URL \url{http://proceedings.mlr.press/v139/radford21a.html}.

\bibitem[Ramos et~al.(2023)Ramos, Michtavy, Porosoff, and White]{ramos2023bayesian}
Mayk~Caldas Ramos, Shane~S Michtavy, Marc~D Porosoff, and Andrew~D White.
\newblock Bayesian optimization of catalysts with in-context learning.
\newblock \emph{arXiv preprint arXiv:2304.05341}, 2023.

\bibitem[Reimers \& Gurevych(2019)Reimers and Gurevych]{reimers2019sentence}
Nils Reimers and Iryna Gurevych.
\newblock Sentence-bert: Sentence embeddings using siamese bert-networks.
\newblock \emph{arXiv preprint arXiv:1908.10084}, 2019.

\bibitem[Rena et~al.(2013)Rena, Pearson, and Sakamoto]{rena2013molecular}
Graham Rena, Ewan~R Pearson, and Kei Sakamoto.
\newblock Molecular mechanism of action of metformin: old or new insights?
\newblock \emph{Diabetologia}, 56:\penalty0 1898--1906, 2013.

\bibitem[Rozemberczki et~al.(2022{\natexlab{a}})Rozemberczki, Gogleva, Nilsson, Edwards, Nikolov, and Papa]{rozemberczki2022moomin}
Benedek Rozemberczki, Anna Gogleva, Sebastian Nilsson, Gavin Edwards, Andriy Nikolov, and Eliseo Papa.
\newblock Moomin: Deep molecular omics network for anti-cancer drug combination therapy.
\newblock In \emph{Proceedings of the 31st ACM International Conference on Information \& Knowledge Management}, pp.\  3472--3483, 2022{\natexlab{a}}.

\bibitem[Rozemberczki et~al.(2022{\natexlab{b}})Rozemberczki, Hoyt, Gogleva, Grabowski, Karis, Lamov, Nikolov, Nilsson, Ughetto, Wang, et~al.]{rozemberczki2022chemicalx}
Benedek Rozemberczki, Charles~Tapley Hoyt, Anna Gogleva, Piotr Grabowski, Klas Karis, Andrej Lamov, Andriy Nikolov, Sebastian Nilsson, Michael Ughetto, Yu~Wang, et~al.
\newblock Chemicalx: A deep learning library for drug pair scoring.
\newblock In \emph{Proceedings of the 28th ACM SIGKDD Conference on Knowledge Discovery and Data Mining}, pp.\  3819--3828, 2022{\natexlab{b}}.

\bibitem[Safavi et~al.(2022)Safavi, Downey, and Hope]{safavi2022cascader}
Tara Safavi, Doug Downey, and Tom Hope.
\newblock Cascader: Cross-modal cascading for knowledge graph link prediction.
\newblock \emph{arXiv preprint arXiv:2205.08012}, 2022.

\bibitem[Sanchez-Lengeling \& Aspuru-Guzik(2018)Sanchez-Lengeling and Aspuru-Guzik]{sanchez2018inverse}
Benjamin Sanchez-Lengeling and Al{\'a}n Aspuru-Guzik.
\newblock Inverse molecular design using machine learning: Generative models for matter engineering.
\newblock \emph{Science}, 361\penalty0 (6400):\penalty0 360--365, 2018.

\bibitem[Scherer et~al.(2022)Scherer, Li{\`o}, and Jamnik]{scherer2022distributed}
Paul Scherer, Pietro Li{\`o}, and Mateja Jamnik.
\newblock Distributed representations of graphs for drug pair scoring.
\newblock \emph{arXiv preprint arXiv:2209.09383}, 2022.

\bibitem[Schwaller et~al.(2021)Schwaller, Probst, Vaucher, Nair, Kreutter, Laino, and Reymond]{schwaller2021mapping}
Philippe Schwaller, Daniel Probst, Alain~C Vaucher, Vishnu~H Nair, David Kreutter, Teodoro Laino, and Jean-Louis Reymond.
\newblock Mapping the space of chemical reactions using attention-based neural networks.
\newblock \emph{Nature Machine Intelligence}, 3\penalty0 (2):\penalty0 144--152, 2021.

\bibitem[Seidl et~al.(2023)Seidl, Vall, Hochreiter, and Klambauer]{seidl2023enhancing}
Philipp Seidl, Andreu Vall, Sepp Hochreiter, and G{\"u}nter Klambauer.
\newblock Enhancing activity prediction models in drug discovery with the ability to understand human language.
\newblock \emph{arXiv preprint arXiv:2303.03363}, 2023.

\bibitem[Singh et~al.(2022)Singh, D'Arcy, Cohan, Downey, and Feldman]{singh2022scirepeval}
Amanpreet Singh, Mike D'Arcy, Arman Cohan, Doug Downey, and Sergey Feldman.
\newblock Scirepeval: A multi-format benchmark for scientific document representations.
\newblock \emph{arXiv preprint arXiv:2211.13308}, 2022.

\bibitem[Snell et~al.(2017)Snell, Swersky, and Zemel]{snell2017prototypical}
Jake Snell, Kevin Swersky, and Richard Zemel.
\newblock Prototypical networks for few-shot learning.
\newblock \emph{Advances in neural information processing systems}, 30, 2017.

\bibitem[Stahl(2020)]{stahl2020prescriber}
Stephen~M Stahl.
\newblock \emph{Prescriber's guide: Stahl's essential psychopharmacology}.
\newblock Cambridge University Press, 2020.

\bibitem[Stankevicius et~al.(2016)Stankevicius, Vasauskas, Noreikiene, Kuodyte, Valius, and Suziedelis]{stankevicius2016extracellular}
Vaidotas Stankevicius, Gintautas Vasauskas, Rimante Noreikiene, Karolina Kuodyte, Mindaugas Valius, and Kestutis Suziedelis.
\newblock Extracellular matrix-dependent pathways in colorectal cancer cell lines reveal potential targets for anticancer therapies.
\newblock \emph{Anticancer Research}, 36\penalty0 (9):\penalty0 4559--4567, 2016.

\bibitem[Su et~al.(2022)Su, Du, Yang, Zhou, Li, Rao, Sun, Lu, and Wen]{su2022molecular}
Bing Su, Dazhao Du, Zhao Yang, Yujie Zhou, Jiangmeng Li, Anyi Rao, Hao Sun, Zhiwu Lu, and Ji-Rong Wen.
\newblock A molecular multimodal foundation model associating molecule graphs with natural language.
\newblock \emph{arXiv preprint arXiv:2209.05481}, 2022.

\bibitem[Sun et~al.(2020)Sun, Wang, Elemento, and Zhou]{sun2020structure}
Mengying Sun, Fei Wang, Olivier Elemento, and Jiayu Zhou.
\newblock Structure-based drug-drug interaction detection via expressive graph convolutional networks and deep sets (student abstract).
\newblock In \emph{Proceedings of the AAAI Conference on Artificial Intelligence}, volume~34, pp.\  13927--13928, 2020.

\bibitem[Tate et~al.(2019)Tate, Bamford, Jubb, Sondka, Beare, Bindal, Boutselakis, Cole, Creatore, Dawson, et~al.]{tate2019cosmic}
John~G Tate, Sally Bamford, Harry~C Jubb, Zbyslaw Sondka, David~M Beare, Nidhi Bindal, Harry Boutselakis, Charlotte~G Cole, Celestino Creatore, Elisabeth Dawson, et~al.
\newblock Cosmic: the catalogue of somatic mutations in cancer.
\newblock \emph{Nucleic acids research}, 47\penalty0 (D1):\penalty0 D941--D947, 2019.

\bibitem[Tunstall et~al.(2022)Tunstall, Reimers, Jo, Bates, Korat, Wasserblat, and Pereg]{tunstall2022efficient}
Lewis Tunstall, Nils Reimers, Unso Eun~Seo Jo, Luke Bates, Daniel Korat, Moshe Wasserblat, and Oren Pereg.
\newblock Efficient few-shot learning without prompts.
\newblock \emph{arXiv preprint arXiv:2209.11055}, 2022.

\bibitem[Tysinger et~al.(2023)Tysinger, Rai, and Sinitskiy]{tysinger2023can}
Emma~P Tysinger, Brajesh~K Rai, and Anton~V Sinitskiy.
\newblock Can we quickly learn to “translate” bioactive molecules with transformer models?
\newblock \emph{Journal of Chemical Information and Modeling}, 63\penalty0 (6):\penalty0 1734--1744, 2023.

\bibitem[Vall et~al.(2021)Vall, Hochreiter, and Klambauer]{vall2021bioassayclr}
Andreu Vall, Sepp Hochreiter, and G{\"u}nter Klambauer.
\newblock Bioassayclr: Prediction of biological activity for novel bioassays based on rich textual descriptions.
\newblock In \emph{ELLIS ML4Molecules workshop}, 2021.

\bibitem[Vaucher et~al.(2021)Vaucher, Schwaller, Geluykens, Nair, Iuliano, and Laino]{vaucher2021inferring}
Alain~C Vaucher, Philippe Schwaller, Joppe Geluykens, Vishnu~H Nair, Anna Iuliano, and Teodoro Laino.
\newblock Inferring experimental procedures from text-based representations of chemical reactions.
\newblock \emph{Nature communications}, 12\penalty0 (1):\penalty0 2573, 2021.

\bibitem[Veli{\v{c}}kovi{\'c} et~al.(2017)Veli{\v{c}}kovi{\'c}, Cucurull, Casanova, Romero, Lio, and Bengio]{velivckovic2017graph}
Petar Veli{\v{c}}kovi{\'c}, Guillem Cucurull, Arantxa Casanova, Adriana Romero, Pietro Lio, and Yoshua Bengio.
\newblock Graph attention networks.
\newblock \emph{arXiv preprint arXiv:1710.10903}, 2017.

\bibitem[von Oswald et~al.(2022)von Oswald, Niklasson, Randazzo, Sacramento, Mordvintsev, Zhmoginov, and Vladymyrov]{von2022transformers}
Johannes von Oswald, Eyvind Niklasson, Ettore Randazzo, Jo{\~a}o Sacramento, Alexander Mordvintsev, Andrey Zhmoginov, and Max Vladymyrov.
\newblock Transformers learn in-context by gradient descent.
\newblock \emph{arXiv preprint arXiv:2212.07677}, 2022.

\bibitem[Wang et~al.(2022)Wang, Liu, Shen, Deng, and Liu]{wang2022deepdds}
Jinxian Wang, Xuejun Liu, Siyuan Shen, Lei Deng, and Hui Liu.
\newblock Deepdds: deep graph neural network with attention mechanism to predict synergistic drug combinations.
\newblock \emph{Briefings in Bioinformatics}, 23\penalty0 (1):\penalty0 bbab390, 2022.

\bibitem[Weininger(1988)]{weininger1988smiles}
David Weininger.
\newblock Smiles, a chemical language and information system. 1. introduction to methodology and encoding rules.
\newblock \emph{Journal of chemical information and computer sciences}, 28\penalty0 (1):\penalty0 31--36, 1988.

\bibitem[Weininger et~al.(1989)Weininger, Weininger, and Weininger]{weininger1989smiles}
David Weininger, Arthur Weininger, and Joseph~L Weininger.
\newblock Smiles. 2. algorithm for generation of unique smiles notation.
\newblock \emph{Journal of chemical information and computer sciences}, 29\penalty0 (2):\penalty0 97--101, 1989.

\bibitem[White et~al.(2022)White, Hocky, Gandhi, Ansari, Cox, Wellawatte, Sasmal, Yang, Liu, Singh, et~al.]{white2022large}
Andrew~D White, Glen~M Hocky, Heta~A Gandhi, Mehrad Ansari, Sam Cox, Geemi~P Wellawatte, Subarna Sasmal, Ziyue Yang, Kangxin Liu, Yuvraj Singh, et~al.
\newblock Do large language models know chemistry?
\newblock \emph{ChemRxiv preprint}, 2022.

\bibitem[White et~al.(2023)White, Hocky, Gandhi, Ansari, Cox, Wellawatte, Sasmal, Yang, Liu, Singh, et~al.]{white2023assessment}
Andrew~D White, Glen~M Hocky, Heta~A Gandhi, Mehrad Ansari, Sam Cox, Geemi~P Wellawatte, Subarna Sasmal, Ziyue Yang, Kangxin Liu, Yuvraj Singh, et~al.
\newblock Assessment of chemistry knowledge in large language models that generate code.
\newblock \emph{Digital Discovery}, 2\penalty0 (2):\penalty0 368--376, 2023.

\bibitem[Xu \& Wang(2022)Xu and Wang]{xu2022protranslator}
Hanwen Xu and Sheng Wang.
\newblock Protranslator: zero-shot protein function prediction using textual description.
\newblock In \emph{Research in Computational Molecular Biology: 26th Annual International Conference, RECOMB 2022, San Diego, CA, USA, May 22--25, 2022, Proceedings}, pp.\  279--294. Springer, 2022.

\bibitem[Xu et~al.(2023{\natexlab{a}})Xu, Woicik, Poon, Altman, and Wang]{xu2023multilingual}
Hanwen Xu, Addie Woicik, Hoifung Poon, Russ~B Altman, and Sheng Wang.
\newblock Multilingual translation for zero-shot biomedical classification using biotranslator.
\newblock \emph{Nature Communications}, 14\penalty0 (1):\penalty0 738, 2023{\natexlab{a}}.

\bibitem[Xu et~al.(2023{\natexlab{b}})Xu, Yuan, Miret, and Tang]{xu2023protst}
Minghao Xu, Xinyu Yuan, Santiago Miret, and Jian Tang.
\newblock Protst: Multi-modality learning of protein sequences and biomedical texts.
\newblock \emph{arXiv preprint arXiv:2301.12040}, 2023{\natexlab{b}}.

\bibitem[Xu et~al.(2019)Xu, Wang, Chen, Tao, and Zhao]{xu2019mr}
Nuo Xu, Pinghui Wang, Long Chen, Jing Tao, and Junzhou Zhao.
\newblock Mr-gnn: Multi-resolution and dual graph neural network for predicting structured entity interactions.
\newblock \emph{arXiv preprint arXiv:1905.09558}, 2019.

\bibitem[Yang et~al.(2023)Yang, Woicik, Poon, and Wang]{yang2023bliam}
Cai Yang, Addie Woicik, Hoifung Poon, and Sheng Wang.
\newblock Bliam: Literature-based data synthesis for synergistic drug combination prediction.
\newblock \emph{arXiv preprint arXiv:2302.06860}, 2023.

\bibitem[Yang et~al.(2021)Yang, Xu, Wu, Chu, and Zhang]{yang2021graphsynergy}
Jiannan Yang, Zhongzhi Xu, William Ka~Kei Wu, Qian Chu, and Qingpeng Zhang.
\newblock Graphsynergy: a network-inspired deep learning model for anticancer drug combination prediction.
\newblock \emph{Journal of the American Medical Informatics Association}, 28\penalty0 (11):\penalty0 2336--2345, 2021.

\bibitem[Yang et~al.(2020)Yang, Jaaks, Dry, Garnett, Menden, and Saez-Rodriguez]{yang2020stratification}
Mi~Yang, Patricia Jaaks, Jonathan Dry, Mathew Garnett, Michael~P Menden, and Julio Saez-Rodriguez.
\newblock Stratification and prediction of drug synergy based on target functional similarity.
\newblock \emph{npj Systems Biology and Applications}, 6\penalty0 (1):\penalty0 16, 2020.

\bibitem[Yao et~al.(2019)Yao, Mao, and Luo]{yao2019kg}
Liang Yao, Chengsheng Mao, and Yuan Luo.
\newblock Kg-bert: Bert for knowledge graph completion.
\newblock \emph{arXiv preprint arXiv:1909.03193}, 2019.

\bibitem[Yasunaga et~al.(2022)Yasunaga, Leskovec, and Liang]{yasunaga2022linkbert}
Michihiro Yasunaga, Jure Leskovec, and Percy Liang.
\newblock Linkbert: Pretraining language models with document links.
\newblock \emph{arXiv preprint arXiv:2203.15827}, 2022.

\bibitem[Youn \& Tagkopoulos(2022)Youn and Tagkopoulos]{youn2022kglm}
Jason Youn and Ilias Tagkopoulos.
\newblock Kglm: Integrating knowledge graph structure in language models for link prediction.
\newblock \emph{arXiv preprint arXiv:2211.02744}, 2022.

\bibitem[Zagidullin et~al.(2019)Zagidullin, Aldahdooh, Zheng, Wang, Wang, Saad, Malyutina, Jafari, Tanoli, Pessia, et~al.]{zagidullin2019drugcomb}
Bulat Zagidullin, Jehad Aldahdooh, Shuyu Zheng, Wenyu Wang, Yinyin Wang, Joseph Saad, Alina Malyutina, Mohieddin Jafari, Ziaurrehman Tanoli, Alberto Pessia, et~al.
\newblock Drugcomb: an integrative cancer drug combination data portal.
\newblock \emph{Nucleic acids research}, 47\penalty0 (W1):\penalty0 W43--W51, 2019.

\bibitem[Zapata et~al.(2020)Zapata, Hansten, Panic, Horn, Boyce, Gephart, Subbian, Romero, and Malone]{zapata2020risk}
Lorenzo~Villa Zapata, Philip~D Hansten, Jennifer Panic, John~R Horn, Richard~D Boyce, Sheila Gephart, Vignesh Subbian, Andrew Romero, and Daniel~C Malone.
\newblock Risk of bleeding with exposure to warfarin and nonsteroidal anti-inflammatory drugs: a systematic review and meta-analysis.
\newblock \emph{Thrombosis and haemostasis}, 120\penalty0 (07):\penalty0 1066--1074, 2020.

\bibitem[Zeng et~al.(2022)Zeng, Yao, Liu, and Sun]{zeng2022deep}
Zheni Zeng, Yuan Yao, Zhiyuan Liu, and Maosong Sun.
\newblock A deep-learning system bridging molecule structure and biomedical text with comprehension comparable to human professionals.
\newblock \emph{Nature communications}, 13\penalty0 (1):\penalty0 862, 2022.

\bibitem[Zhao et~al.(2023)Zhao, Zhou, Cao, Zhang, and Chen]{zhao2023adversarial}
Wenyu Zhao, Dong Zhou, Buqing Cao, Kai Zhang, and Jinjun Chen.
\newblock Adversarial modality alignment network for cross-modal molecule retrieval.
\newblock \emph{IEEE Transactions on Artificial Intelligence}, 2023.

\bibitem[Zheng et~al.(2021)Zheng, Aldahdooh, Shadbahr, Wang, Aldahdooh, Bao, Wang, and Tang]{zheng2021drugcomb}
Shuyu Zheng, Jehad Aldahdooh, Tolou Shadbahr, Yinyin Wang, Dalal Aldahdooh, Jie Bao, Wenyu Wang, and Jing Tang.
\newblock Drugcomb update: a more comprehensive drug sensitivity data repository and analysis portal.
\newblock \emph{Nucleic acids research}, 49\penalty0 (W1):\penalty0 W174--W184, 2021.

\end{thebibliography}
